\documentclass[lettersize,journal]{IEEEtran}
\usepackage{amsmath,amsfonts}
\usepackage{algorithmic}
\usepackage{algorithm}
\usepackage{array}
\usepackage[caption=false,font=normalsize,labelfont=sf,textfont=sf]{subfig}
\usepackage{textcomp}
\usepackage{stfloats}
\usepackage{url}
\usepackage{physics} 
\usepackage{bm}
\usepackage{verbatim}
\usepackage{graphicx}
\usepackage{cite}
\usepackage{booktabs}
\usepackage{multirow,booktabs,graphicx,caption,subcaption,xcolor}
\usepackage{subcaption}
\usepackage{amsmath,amsfonts}
\usepackage{algorithmic}
\usepackage{algorithm}
\usepackage{array}
\usepackage[caption=false,font=normalsize,labelfont=sf,textfont=sf]{subfig}
\usepackage{textcomp}
\usepackage{stfloats}
\usepackage{url}
\usepackage{verbatim}
\usepackage{graphicx}
\usepackage{cite}
\usepackage{multirow,booktabs}
\usepackage{multirow} 
\usepackage{color}
\usepackage{xcolor}
\usepackage{color}
\usepackage{colortbl}
\usepackage[colorlinks,linkcolor=blue]{hyperref}
\usepackage{amsmath,amssymb,amsfonts}
\usepackage{algorithmic}
\usepackage{graphicx}
\usepackage{comment}

\begin{document}

\title{SPIDER: Multi-Layer Semantic Token Pruning and Adaptive Sub-Layer Skipping in Multimodal Large Language Models}
\author{Tianxiang Chen, Zhentao Tan, Zi Ye, Yue Wu, Xiaobing Tu, Jinkui Ren, Xiantao Zhang, Tao Gong*, Qi Chu, Nenghai Yu, Xipeng Qiu, Jieping Ye, \IEEEmembership{Fellow, IEEE}
\thanks{Tianxiang Chen, Xiaobing Tu, Jinkui Ren, and Xiantao Zhang are with the End-User Intelligent Computing BU, Alibaba Cloud (e-mail: txchen@ustc.edu.cn, xiaobing.tuxiaobin@alibaba-inc.com, guancheng.rjk@alibaba-inc.com, xiantao.zxt@alibaba-inc.com). Tao Gong, Qi Chu and Nenghai Yu are with the School of Cyber Science and Technology, University of Science and Technology of China (e-mail: tgong@ustc.edu.cn, qchu@ustc.edu.cn, ynh@ustc.edu.cn). Zi Ye is with the Department of Computer Science, Maynooth University (e-mail: zi.ye@mu.ie). Xipeng Qiu is with Fudan University (e-mail: xpqiu@fudan.edu.cn). Zhentao Tan, Yue Wu, and Jieping Ye are with Alibaba Group (e-mail: zhentaotan5@gmail.com, matthew.wy@alibaba-inc.com, yejieping.ye@alibaba-inc.com). \\ *Tao Gong is the corresponding author.}}

\markboth{Journal of \LaTeX\ Class Files,~Vol.~14, No.~8, August~2021}%
{Shell \MakeLowercase{\textit{et al.}}: A Sample Article Using IEEEtran.cls for IEEE Journals}


\maketitle

\begin{abstract}

Multimodal Large Language Models face significant efficiency challenges that stem from two distinct yet coupled sources: data redundancy and computational redundancy. While most methods focus on data redundancy by pruning visual tokens from the output of the visual encoder or computing redundancy in LLM decoders using blockwise importance, the finer-grained inter-layer representation shifts and the distribution differences within the layers themselves have not been fully explored. In this work, we comprehensively investigate this dual-level inefficiency. We posit that intermediate layer tokens from vision encoders should be considered for effective visual token pruning, as semantic focus shifts across layers, with middle-layer tokens capturing more detailed object-centric information that deeper layers may abstract away. Furthermore, we reveal the differential contributions of Attention and FFNs across distinct LLM decoder layers. Building upon these discoveries, we propose \textbf{SPIDER}, a training-free framework that integrates multi-layer \underline{\textbf{S}}emantic visual token \underline{\textbf{P}}run\underline{\textbf{I}}ng with an a\underline{\textbf{D}}aptive sub-lay\underline{\textbf{ER}} skipping mechanism. Experimental evaluations demonstrate that SPIDER consistently maintains strong performance across various MLLM architectures and reduction ratios. For instance, on LLaVA-NeXT-7B, SPIDER reduces FLOPs by $79\%$ while maintaining 96$\%$ of the baseline performance.


\end{abstract}


\section{Introduction}
\IEEEPARstart{M}{ultimodal} large language models (MLLMs) integrate visual information with powerful large language models (LLMs), achieving strong performances on various complex tasks, such as image understanding \cite{bai2025qwen2}, video understanding \cite{lin2024video}, and visual reasoning \cite{zhangLLaVA}. However, this integration introduces significant overhead, stemming from both data redundancy in the form of lengthy visual token sequences and computational redundancy within the large-scale LLM backbone. While visual token pruning has become a dominant strategy to tackle data redundancy \cite{chen2024image, zhangsparsevlm, zhang2024beyond}, the computational redundancy in processing the remaining tokens through the LLM decoder has received insufficient attention. In this work, we argue that data redundancy and computational redundancy can be combined, and each requires a more fine-grained pruning strategy. 

\begin{figure*}
    \centering \includegraphics[width=0.99\textwidth,height=0.43\textwidth]{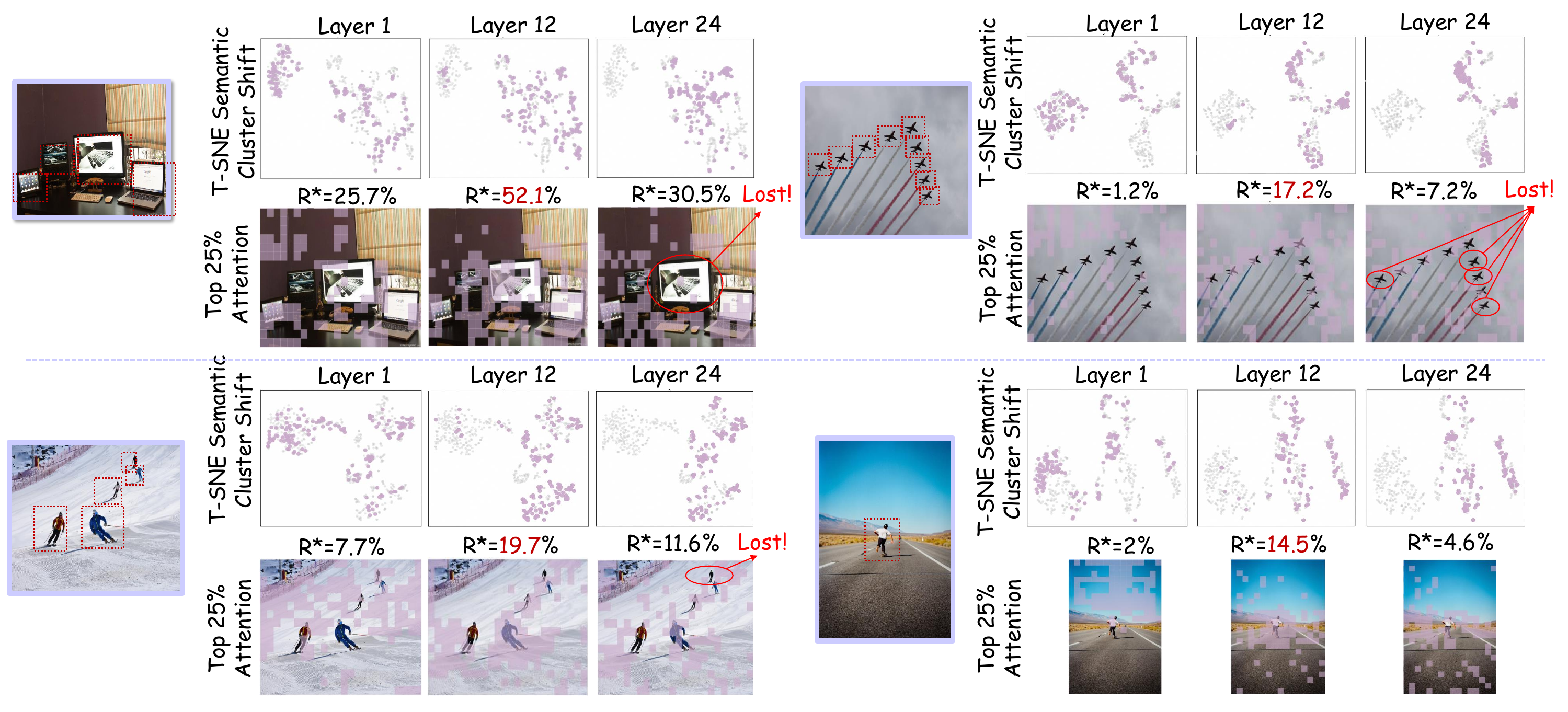}
    \caption{Visualizations to show the importance of mid-layer features for fine-grained token pruning. We highlight the top 25$\%$ most attentive tokens (in purple) from various layers of the CLIP vision encoder, together with their t-SNE embeddings. Mid-layer tokens exhibit a higher key object coverage ratio ($R*$), indicating a stronger focus on crucial object-centric details compared to deeper layers, which tend to capture broader global semantics. Averaged over 100 sampled evaluation images, $R^*$ peaks at $29.65\%$ at layer~12, compared to $11.28\%$ at layer~1 and $15.83\%$ at layer~24.}
    \label{shift}
\end{figure*}

First, on the data redundancy front, most state-of-the-art token pruners \cite{zhang2024beyond, zhangsparsevlm, chen2024image} rely exclusively on the feature maps from the final layer of the vision encoder to compute token importance. However, our empirical analysis suggests that this ``last-layer-only'' approach suffers from a significant semantic focus shift. As illustrated in Fig. \ref{shift}, while the deep layers of a Vision Transformer (ViT) excel at capturing abstract, global context, the middle layers often retain superior object-centric details essential for fine-grained tasks like counting or precise localization. By discarding middle-layer insights, present pruning methods risk eliminating critical visual fragments that are semantically ``diluted'' in the final layer but vital for accurate reasoning.

Second, on the computational redundancy front, existing acceleration methods primarily focus on coarse-grained block-skipping \cite{lawson2025learning, csordasneural}. These approaches indiscriminately bypass entire LLM decoder blocks, ignoring the internal heterogeneity of the decoder. Our diagnostic experiments reveal a functional divergence between sub-layers: the Attention and Feed-forward Network (FFN) components contribute unequally to the refinement of visual tokens across the decoding stages. This finding motivates a more adaptive and fine-grained skipping strategy than treating blocks as monolithic units.

Based on these findings, we propose SPIDER, a novel, training-free framework to improve MLLM inference efficiency. It is built on two core mechanisms: \textbf{Multi-layer semantic token pruning (MSV-Prune)}: This strategy uses tokens from both deep and middle layers of the visual encoder for semantic clustering and similarity computation. \textbf{Adaptive sub-layer skipping (ASL-Skip)}: This strategy accumulates a skippability score to determine which retained visual tokens should perform layer skipping and at which layer to skip, and an offline sub-layer contribution score ($\mathrm{SLC}$) to decide which specific sub-layer (Attention or FFN) to skip in subsequent LLM decoder layers for skipped tokens. 

A key perspective behind SPIDER is that the utility of a visual token changes with depth across the full multimodal inference pipeline. Accordingly, MSV-Prune determines which tokens should enter the decoder at all, while ASL-Skip determines how much computation each retained token still needs as decoding progresses. SPIDER is therefore a unified utility-allocation framework rather than a simple combination of two independent acceleration modules. Our contributions are summarized as follows:
\begin{itemize}
\item We explore the semantic focus shift across vision encoder layers, and propose multi-layer semantic token pruning, considering middle and deep layer tokens.
\item We quantify the fine-grained redundancy in attention and FFN sub-layers of MLLM decoders, and propose adaptive sub-layer skipping to decide whether a retained visual token should skip, when to skip, and skip which part of the layers.
\item We propose SPIDER, a unified training-free framework that 
jointly addresses encoder-side token redundancy and decoder-side computational redundancy, significantly reducing inference cost across various MLLM architectures while maintaining comparable performance.
\end{itemize}

\section{Related Works}
\label{gen_inst}

\subsection{Multimodal Large Language Models (MLLMs)}
MLLMs have emerged as a dominant paradigm for unified vision–language understanding and generation, leveraging the strong priors of large language models (LLMs) to interpret and reason over visual inputs \cite{comanici2025gemini, liu2024improved}. Pioneering architectures such as LLaVA \cite{liu2024improved}, BLIP-2 \cite{li2023blip}, and MiniGPT-4 \cite{zhu2023minigpt} typically employ a frozen vision encoder (e.g., CLIP ViT) to extract image features, which are then projected into the LLM’s textual embedding space via a trainable connector. This design enables zero-shot or few-shot multimodal inference through natural language prompts, achieving remarkable performance across diverse tasks—from visual question answering to image captioning.

Beyond general-purpose benchmarks, MLLMs have been successfully specialized for domain-critical applications. For instance, EmoVerse \cite{li2024emoverse} is specifically designed for affective computing, integrating emotion-aware visual reasoning through multi-task training to jointly handle sentiment analysis, emotion recognition, facial expression interpretation, and emotion cause inference. While MedTVT \cite{zhang2025medtvt} integrates clinical multimodal data and medical knowledge graphs to enable interpretable, evidence-based multi-disease diagnosis through chain-of-evidence reasoning. These advances underscore the potential of MLLMs in high-stakes scenarios where accuracy, reliability, and contextual grounding are paramount.

However, the computational overhead of MLLMs remains a fundamental barrier to real-world deployment. High-resolution images often yield thousands of visual tokens after patch embedding, drastically inflating sequence length and memory consumption during LLM decoding \cite{zhang2024LLaVA, zhang2024beyond}. This issue is exacerbated in video understanding \cite{maaz2023video, zhang2023video, lin2024video}, where temporal redundancy across frames compounds token explosion. Consequently, even state-of-the-art MLLMs struggle with latency and scalability in edge or interactive settings, necessitating principled efficiency mechanisms that preserve semantic fidelity without retraining.

\begin{figure}
    \centering \includegraphics[width=0.49\textwidth,height=0.32\textwidth]{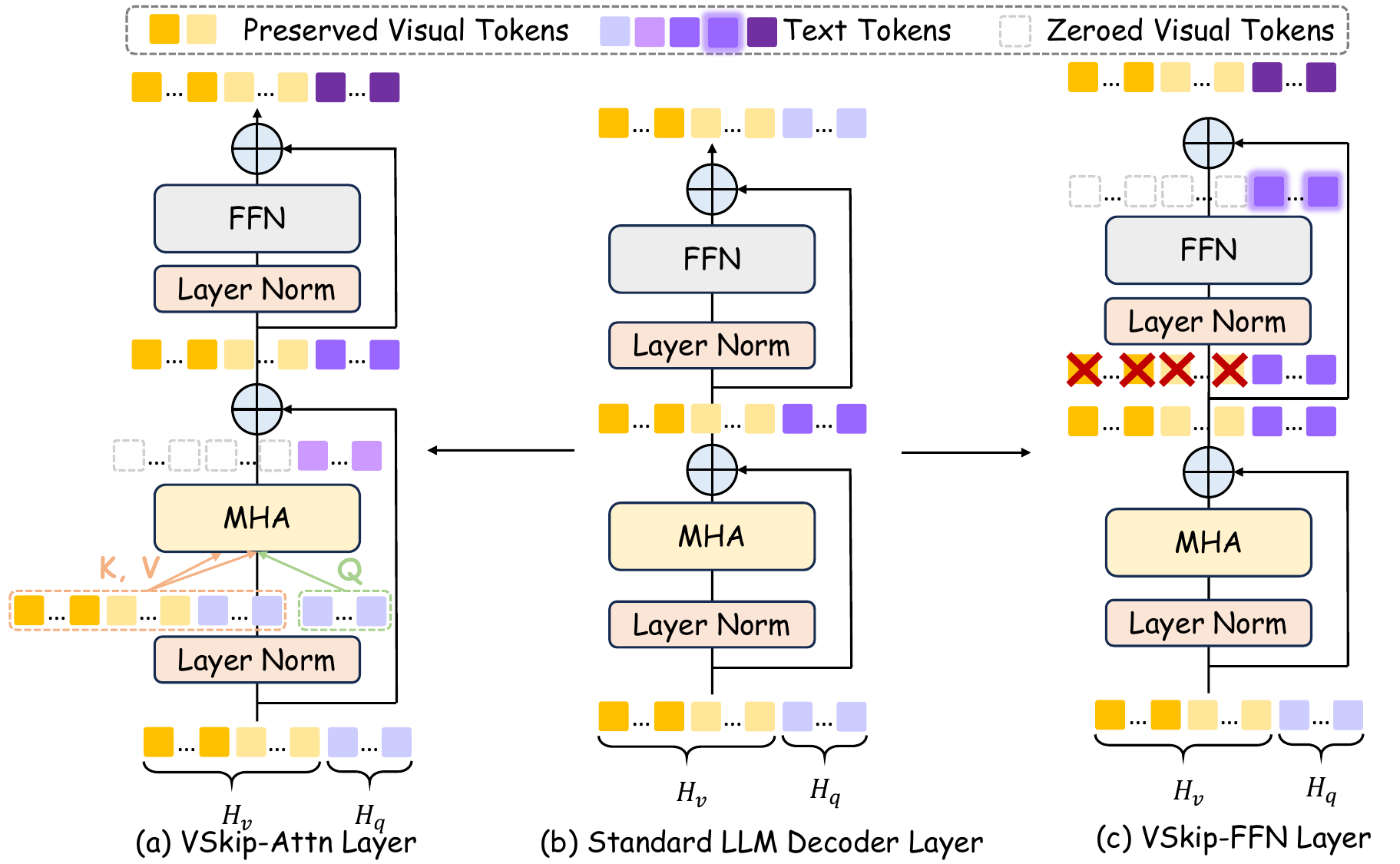}
    \caption{Visualization of sub-layer skipping, where we visualize either (a) skipping attention sub-layer module or (c) FFN sub-layer module. Sub-layer skipping is achieved via replacing the LLM decoder block with corresponding sparse blocks, so that visual tokens are blocked in attention or FFN modules.
    }\label{layer}
\end{figure}

\subsection{Reducing Redundancy in MLLMs}

While MLLMs have achieved remarkable success, their huge computational cost hinders scalable deployment. Existing inference optimizations primarily fall into two categories: token pruning and layer skipping.

Token pruning has gained wide attention due to the high redundancy of visual tokens, which dominate input sequences. Some efforts focused on compressing visual tokens into compact representations \cite{chen2024efficient, li2024llama, shang2024LLaVA}, but requiring additional training. Other training-free methods leverage text-visual attention within the LLM to identify less important visual tokens \cite{chen2024image, zhangsparsevlm, xing2024pyramiddrop}. However, studies like Vispruner \cite{zhang2024beyond}, VisionZip \cite{yang2025visionzip}, VTC-CLS \cite{wang2024cls}, and HiPrune \cite{liu2025hiprune} argue against the sole reliance on text-visual attention due to positional bias, advocating for visual cues. A common limitation across most token pruning methods is their focus on pruning from the final output, neglecting the multi-layer semantic information inherent in the vision encoder.

Layer skipping addresses the inherent layer redundancy in MLLMs, as observed by ShortV \cite{yuan2025shortv}. While some layer skipping approaches involve additional training \cite{zeng2025skip, suo2024pruning, elhoushi2024layerskip}, our focus remains on training-free methods. ShortV \cite{yuan2025shortv}, for instance, identifies and replaces less effective layers with sparse versions where visual tokens remain frozen. However, these methods are not fine-grained enough since they often aggressively replace entire LLM layers for all visual tokens, overlooking token and sub-layer importance variance . 

Different from previous works, we use the semantic information from middle and deep layer tokens for token pruning. Besides, we systematically analyze the contribution difference of attention and FFN modules in the LLM decoder and adopt a more fine-grained sub-layer skipping mechanism.


Unlike prior work that treats token pruning and layer skipping as independent problems, SPIDER addresses them as two phases of a single depth-dependent utility-allocation problem: visual token utility shifts across encoder layers motivate multi-layer pruning before decoding, while the gradual semantic convergence of retained tokens during decoding motivates adaptive sub-layer skipping. Concretely, unlike previous token pruning methods, we explicitly consider the semantic differences between middle and deep layers of the vision encoder, leveraging a multi-layer semantic clustering approach to achieve more comprehensive token pruning. Furthermore, diverging from coarse-grained layer skipping strategies, SPIDER introduces a sub-layer skipping mechanism that adaptively determines whether a retained visual token should skip, when to skip, and which sub-layer to skip. The two modules are thus mutually motivated by the same depth-dependent view of token utility, rather than being independently assembled.

\section{Key Findings}
\label{key_findings}
\subsection{Semantic Shift across Vision Encoder}

We visualize the semantic focus shift and the distribution of the top 25$\%$ attentive tokens across CLIP vision encoder layers in Figure~\ref{shift}. t-SNE embeddings across layers show a gradual shift in semantic clustering, with middle layers bridging distinct visual concepts. Notably, middle-layer tokens focus more on key object regions, yielding higher key object coverage ratios $R^*$ compared to early and deep layers. To quantify this systematically, we compute the average $R^*$ over 100 sampled evaluation images at three representative layers of the 24-layer CLIP ViT: $11.28\%$ at layer~1 (early), $29.65\%$ at layer~12 (middle), and $15.83\%$ at layer~24 (deep), confirming a clear middle-layer peak consistent with the per-example results in Fig.~\ref{shift}. Deep layers, while capturing broader scene regions, risk missing critical object information. This suggests that effective token pruning should leverage both middle- and deep-layer features, not solely the deepest layer.



\subsection{Sub-Layer Redundancy in MLLMs}
To quantify sub-layer redundancy for specific tokens, we introduce two sparse layers: \textbf{VSkip-Attn} and \textbf{VSkip-FFN}. In the VSkip-Attn layer, visual tokens do not function as queries, acting solely as keys and values. In the VSkip-FFN layer, the visual token input to the FFN is directly bypassed.

We propose the \textbf{Sub-Layer Contribution (SLC) score} to quantify a sub-layer's impact on the model's final prediction. The SLC is calculated as the Kullback-Leibler (KL) divergence between the output logits of the original model and a modified one where a specific sub-layer's operation is selectively bypassed. 
Specifically, to measure the contribution, we simulate skipping a sub-layer (e.g., attention or FFN) for a subset of tokens $X$ at a given layer $i$ by preventing their hidden states from being updated by that sub-layer. We then define $\mathbf{SLC}_{i,\mathrm{attn}}^X$ and $\mathbf{SLC}_{i,\mathrm{ffn}}^X$ as the KL divergence when the self-attention or FFN sub-layers are skipped for tokens $X$ in layer $i$, respectively. A lower $\mathbf{SLC}$ value indicates less influence on the final output, making the sub-layer a stronger skip candidate. For each layer, we identify the least impactful module by comparing the two sub-layers after the per-module normalization introduced in Eq.~(5), and replace modules with the corresponding sparse layers in order of decreasing skippability. 

\begin{figure*}
    \centering \includegraphics[width=0.99\textwidth,height=0.36\textwidth]{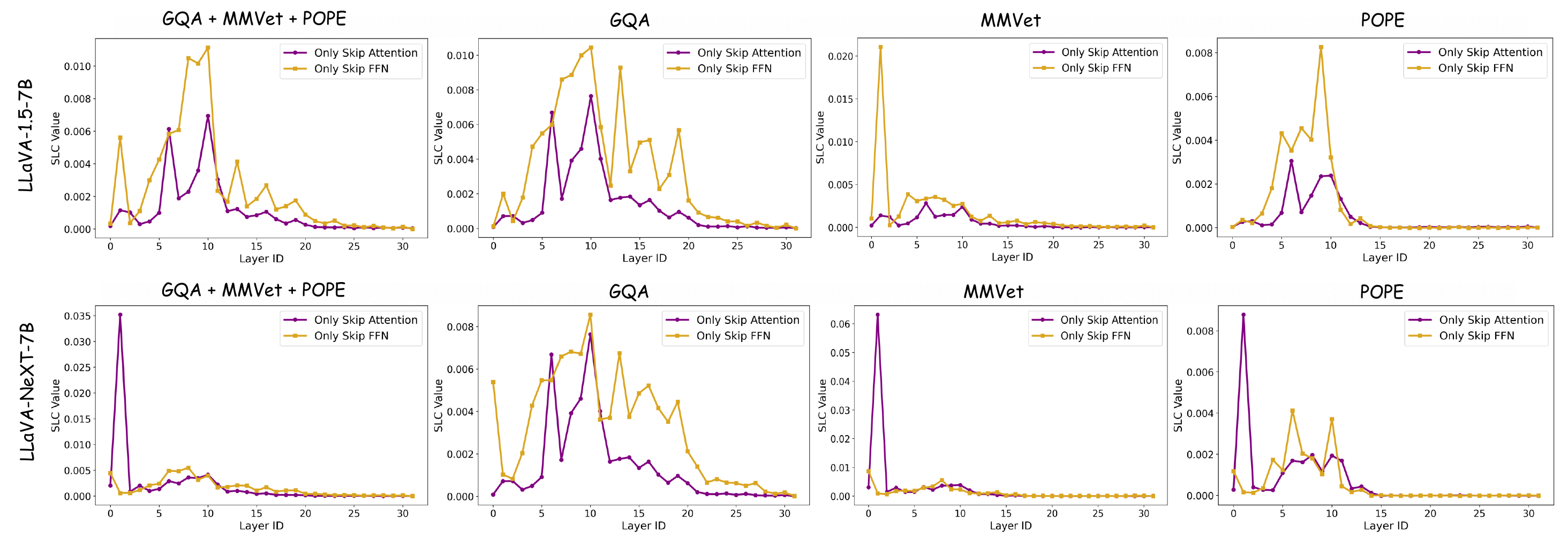}
    \caption{The sub-layer contribution scores ($\mathbf{SLC}$) of LLaVA-1.5-7B and LLaVA-NeXT-7B. Lower $\mathbf{SLC}$ values indicate a weaker influence of the corresponding attention or FFN sub-layer on the specified tokens. Skipping transformations of visual tokens in such low-impact sub-layers yields minimal divergence from the original model’s output distribution, and the $\mathbf{SLC}$ distribution is very different across various benchmarks.
}\label{lc_vis}
\end{figure*}

To ensure robustness, we compute $\mathbf{SLC}$ offline across three diverse benchmarks. As shown in Figure~\ref{lc_vis}, the resulting SLC distributions exhibit significant variance in absolute magnitude across datasets. However, the relative sub-layer rankings remain highly consistent: averaged across benchmark pairs, the top-16 most-skippable attention sub-layers overlap by $12.7/16$ on LLaVA-1.5-7B and $13.3/16$ on LLaVA-NeXT-7B; the corresponding FFN overlaps are $13.0/16$ and $13.0/16$. This rank consistency justifies averaging the SLC across benchmarks to obtain a generalizable backbone-level skipping policy. Another key finding is that $\mathbf{SLC}_{i,\mathrm{attn}}^X$ is generally lower than $\mathbf{SLC}_{i,\mathrm{ffn}}^X$. This suggests that for visual tokens, the attention sub-layer is often more redundant than the FFN, making it a preferable target for skipping.

\section{Method}
\label{headings}
Our training-free framework is illustrated in Figure~\ref{overview}. The input image first undergoes \textbf{multi-layer semantic visual token pruning} to reduce data redundancy, and the retained tokens are then fed to the LLM decoder for \textbf{adaptive sub-layer skipping}. The two components are designed to be coupled rather than independent. MSV-Prune determines which visual tokens should enter the decoder at all by removing semantically redundant tokens before decoding, while ASL-Skip further determines how much computation each retained token still needs as decoding progresses. In this way, SPIDER allocates computation across the full multimodal inference pipeline according to the evolving utility of visual tokens. Details are as follows.


\subsection{Multi-layer Semantic Visual Token Pruning}
The retained visual tokens comprise two subsets: a fraction $r$ of anchor tokens $\mathbf{T}^{\mathrm{anc}}_v$ selected via attention sorting, and complementary tokens  $\mathbf{T}^{\mathrm{cmp}}_v$ selected under the involvement of middle layer tokens into semantic clustering and sorting. Then  $\mathbf{T}^{\mathrm{anc}}_v$ and $\mathbf{T}^{\mathrm{cmp}}_v$ are concatenated in spatial indices order and fused with text tokens $T_q$ for LLM decoding.

\subsubsection{Attention-based anchor tokens}
To address the high redundancy in visual inputs, we first perform a saliency-based token pruning step. Drawing inspiration from findings that visual encoder attention is a reliable indicator of patch importance \cite{zhang2024beyond}, we use the visual encoder's self-attention scores to select a compact set of important tokens. We average the self-attention matrix $\mathbf{A}$ over all heads to obtain $\mathbf{a}_v\in\mathbb{R}^n$, where for CLIP-like encoders $\mathbf{a}_v$ is the \texttt{[CLS]} row, and for encoders without \texttt{[CLS]} it is the mean attention each patch token receives. A dynamic threshold $\tau$ selects anchor tokens to meet a budget of $n \times R \times r$ tokens, where $R \in (0,1)$ is the overall retention ratio and $r \in (0,1)$ is a hyperparameter defining the proportion of anchor tokens in the retained set.
\begin{equation}
\begin{split}
    \tau = \min &\{t \mid |\{a_i^v \geq t\}| \leq n \times R \times r \}, \\
    \mathbf{T}_v^{\mathrm{anc}} &= \{t_i^v \in \mathbf{T}_v \mid a_i^v \geq \tau\}.
\end{split}
\end{equation}


\subsubsection{Multi-layer Semantic clustering-based complementary tokens}

Relying solely on foreground-centric anchor tokens ($\mathbf{T}^{\text{anc}}_v$) risks losing crucial background context. To mitigate this, we propose a coarse-to-fine strategy to select a set of complementary tokens ($\mathbf{T}^{\text{cmp}}_v$). $\mathbf{T}^{\text{cmp}}_v$ are obtained by clustering non-anchor tokens into $K$ groups, computing their multi-layer similarity scores, selecting the $N_k$ lowest-scoring tokens from each cluster, and aggregating them.

\textbf{Coarse-grained Semantic Clustering.} 
A naive redundancy removal on non-anchor tokens is suboptimal, as populous but uniform background regions would exhaust the selection quota, displacing smaller yet unique semantic areas. To address this, we apply K-means clustering to the high-level features $\mathbf{F}_v^L$ of non-anchor tokens, partitioning them into $K$ semantic clusters $\{C_k\}_{k=1}^K$. The total budget for complementary tokens, $N_{\text{cmp}} = n \times R \cdot (1-r)$, is then allocated proportionally to each cluster as a quota $N_k$, ensuring all semantic groups are represented.
\begin{equation}
    N_k = \max\left(1, \text{round}\left(N_{\text{cmp}} \cdot \frac{|C_k|}{\sum_{j=1}^{K} |C_j|}\right)\right)
    \label{eq:quota_allocation}
\end{equation}

\textbf{Fine-grained Multi-view Pruning.} 
Within each semantically homogeneous cluster $C_k$, we select the most informative $N_k$ tokens by leveraging hierarchical features. We observe that middle-layer features ($\mathbf{F}^{M}$) capture fine-grained object-centric details, whereas last-layer features ($\mathbf{F}^{L}$) encode global semantics, as analyzed in Section~\ref{key_findings}. To select a complementary set, we compute a multi-layer similarity score $\mathbf{S}_{ij}$ for any pair of tokens $(i, j)$. This score considers their similarity at both middle and deep layer feature levels (multi-layer intra-cluster similarity), and their similarity to the already-selected anchor tokens:

\begin{equation}
\begin{split}
    &\mathbf{S}_{ij} = \underbrace{\left(\text{sim}(\mathbf{F}_{i}^{L}, \mathbf{F}_{j}^{L}) + \text{sim}(\mathbf{F}_{i}^{M}, \mathbf{F}_{j}^{M})\right)}_{\text{Multi-layer Intra-Cluster Similarity}} + \\
    &\underbrace{\left( \max_{p \in \mathbf{T}_v^{\text{anc}}} \text{sim}(\mathbf{F}_{i}^{L}, \mathbf{F}_p^L) + \max_{p \in \mathbf{T}_v^{\text{anc}}} \text{sim}(\mathbf{F}_{j}^{L}, \mathbf{F}_p^L) \right)}_{\text{Similarity to Anchor Tokens}}
\end{split}
\end{equation}
where $\text{sim}(\cdot, \cdot)$ is the cosine similarity. For each cluster, we compute similarity scores for all tokens, retain the $N_k$ lowest-scoring ones, and aggregate them across clusters to form $\mathbf{T}^{\text{cmp}}_v$.

\subsection{Adaptive Sub-Layer Skipping}
Even after token pruning, significant computational redundancy persists within the MLLM decoder. While prior work has explored coarse-grained layer skipping, this often incurs performance degradation. We observe that sub-layers (i.e., self-attention and FFN) within each decoder block contribute unequally to the final output (Fig.~\ref{lc_vis}). This motivates our fine-grained, \emph{adaptive sub-layer skipping} policy, which dynamically bypasses less critical sub-layers for specific tokens. The decision process is guided by a hybrid mechanism, combining an online, per-token skippability score with an offline, per-layer sub-layer contribution score.

\subsubsection{Online Skippability Score}

To determine \emph{if} and \emph{at which layer} a token should perform sub-layer skipping, we compute an online skippability score $\mathbf{S}_{sa}(i,\ell)$ for each visual token $i$ at each layer $\ell$. It consists of two aspects:

\textbf{1. Intrinsic Information Entropy ($\mathbf{E}_{ii}$):} This measures the semantic uncertainty of a token's hidden state $h_{\text{v}}(i,\ell)$. We compute it as the Shannon entropy of the probability distribution 
$p(i, \ell) = \operatorname{Softmax}(h_v(i, \ell) \cdot W_{\text{unembed}}^{\top})$, 
which results from projecting the hidden state into the vocabulary space $V$ via the language model’s unembedding matrix $W_{\text{unembed}} \in \mathbb{R}^{V\times d}$. Low entropy indicates that the token's hidden state has semantically converged and is unlikely to benefit from further sub-layer updates, making it a candidate for skipping. We normalize the entropy to $[0, 1]$ to get $\mathbf{E}_{ii}(i,\ell)$.  


\textbf{2. Image-Text Correlation Factor ($\mathbf{F}_{itc}$):} This is a common factor \cite{zhangsparsevlm, chen2024image} that assesses a visual token's relevance to the text query. A token weakly correlated with the text context is more skippable. We measure this as the cosine similarity between the visual token's hidden state $h_{\text{v}}(i,\ell)$ and the averaged text context vector $q_{t}(\ell)$. The cosine similarity $\text{sim}(i,\ell)$ is also normalized to $[0, 1]$ to be $\mathbf{F}_{itc}$. We then define a \emph{retention score} $\mathbf{R}(i,\ell) = \mathbf{E}_{ii}(i,\ell) + \mathbf{F}_{itc}(i,\ell)$. The final skippability score is its inverse: 
\begin{equation}
    \mathbf{S}_{sa}(i,\ell) = \text{ReLU}\left( 1 - \mathbf{R}(i,\ell) \right).
\end{equation}
A higher $\mathbf{S}_{sa}$ indicates a stronger signal for skipping. Since both $\mathbf{E}_{ii}$ and $\mathbf{F}_{itc}$ are independently normalized to $[0,1]$, their linear combination is a natural and interpretable form of evidence aggregation that requires no additional learned parameters. The ablation in Table~\ref{ab:ASL-Skip}(c) confirms that both components contribute meaningfully, validating this design.

\begin{algorithm}[t]
\caption{Pseudocode for SPIDER}
\label{alg:spider}
\begin{algorithmic}[1]
\REQUIRE $x$, $\mathcal{E}$, $\mathcal{D}$, $R$, $r$, $T_{\text{skip}}$, $w_1, w_2$
\ENSURE $y$

\STATE // \textbf{Stage 1: Multi-layer Visual Token Pruning}
\STATE Extract mid/last features: $\mathbf{F}^M, \mathbf{F}^L \gets \mathcal{E}(x)$
\STATE Select anchor tokens $\mathbf{T}^{\mathrm{anc}}$ (top-$nRr$ by attention score)
\STATE Cluster non-anchor tokens on $\mathbf{F}^L$; select representatives minimizing multi-layer similarity to anchors and within-cluster redundancy
\STATE $\mathbf{T}_{\text{input}} \gets [\mathbf{T}_q; \text{concat}(\mathbf{T}^{\mathrm{anc}}, \mathbf{T}^{\mathrm{cmp}})]$

\STATE // \textbf{Stage 2: Adaptive Sub-Layer Skipping}
\STATE Precompute $\mathrm{SLC}^{\text{Norm}}_{\text{Attn/FFN}}(\ell)$ and the per-layer skip target $m(\ell) \gets \arg\max_{module} \mathrm{SLC}^{\text{Norm}}_{module}(\ell)$
\STATE $\mathrm{S}_{skip}(i) \gets 0$;\ \ $\mathcal{M} \gets \emptyset$;\ \ $h \gets \mathbf{T}_{\text{input}}$ \quad // $\mathcal{M}$: tokens in skip mode
\FOR{$\ell = 1$ to $L$}
    \IF{$\ell \leq L/2$}
        \STATE // decision window: accumulate evidence, admit new tokens
        \FOR{each visual token $i \notin \mathcal{M}$}
            \STATE Compute $\mathbf{E}_{ii}(i,\ell)$, $\mathbf{F}_{itc}(i,\ell)$, and $\mathbf{S}_{sa}(i,\ell)=\mathrm{ReLU}\!\left(1-\mathbf{E}_{ii}(i,\ell)-\mathbf{F}_{itc}(i,\ell)\right)$
            \STATE $\mathrm{S}_{skip}(i) \mathrel{+}= w_1 \cdot \mathbf{S}_{sa}(i,\ell) + w_2 \cdot \mathrm{SLC}^{\text{Norm}}_{m(\ell)}(\ell)$
            \IF{$\mathrm{S}_{skip}(i) \geq T_{\text{skip}}$}
                \STATE $\mathcal{M} \gets \mathcal{M} \cup \{i\}$;\ \ $\mathrm{target}(i) \gets m(\ell)$ \quad // frozen at admission
            \ENDIF
        \ENDFOR
    \ENDIF
    \STATE $h \gets$ Forward layer $\ell$, bypassing sub-layer $\mathrm{target}(i)$ for every $i \in \mathcal{M}$
\ENDFOR
\STATE $y \gets \mathrm{LMHead}(h)$
\RETURN $y$
\end{algorithmic}
\end{algorithm}

\begin{figure*}
    \centering \includegraphics[width=0.99\textwidth,height=0.25\textwidth]{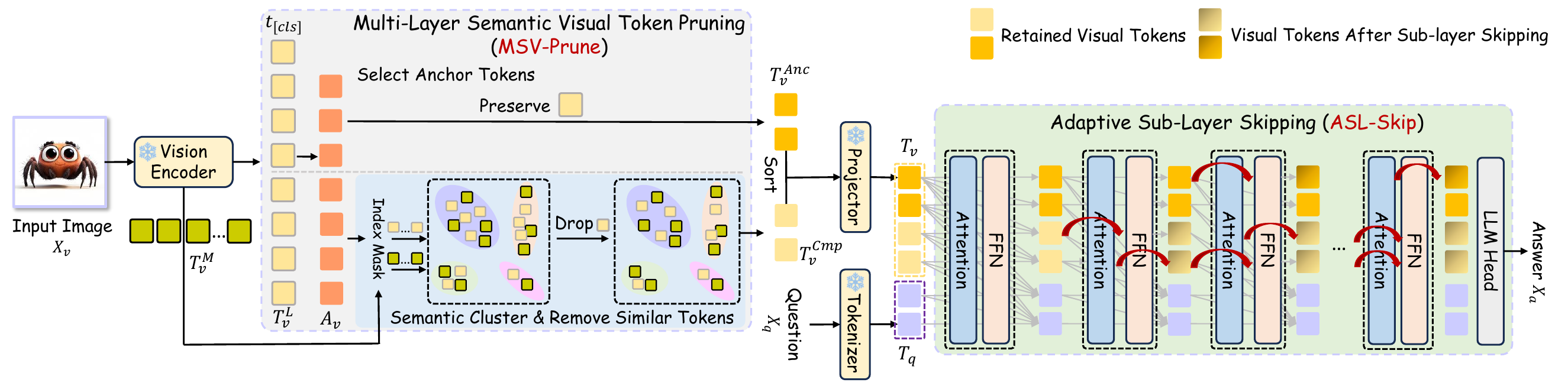}
    \caption{\textbf{Illustration of SPIDER}. We begin by token pruning using the semantic information from the middle and last layers. The retained tokens are fed to the LLM for adaptive sub-layer skipping.}\label{overview}
\end{figure*}

\subsubsection{Offline Sub-Layer Contribution Score}
Once a token is deemed skippable, we must decide \emph{which} sub-layer (attention or FFN) to bypass. To inform this, we pre-compute a Sub-Layer Contribution (SLC) score via offline profiling. For each layer $\ell$, $\mathrm{SLC}_{attn}(\ell)$ and $\mathrm{SLC}_{ffn}(\ell)$ are the average KL-divergence between the original model's output and the output when the respective sub-layer is skipped via sparse layer replacement, evaluated over multiple multimodal benchmarks. A low KL-divergence implies the module is less critical. We normalize these to obtain module-level skippability scores, where a higher score means more skippable:
\begin{equation}
\begin{split}
    &\mathrm{SLC}^{Norm}_{module}(\ell) = 1 - \frac{\mathrm{SLC}_{module}(\ell) - \mathrm{SLC}_{module, min}}      
    {\mathrm{SLC}_{module, max} - \mathrm{SLC}_{module, min} + \epsilon},\\ &\quad module \in \{\text{Attn, FFN}\}.
\end{split}
\end{equation}

At each layer $\ell$, the sub-layer with the higher $\mathrm{SLC}^{Norm}$ score is designated as that layer's skip target $m(\ell)$, and Table~\ref{tab:replaced_layers} lists $m(\ell)$ for every decoder index. Two aspects of this designation are worth making explicit. First, the comparison is made between the two \emph{normalized} scores rather than between the raw KL values, because the attention and FFN families occupy systematically different KL ranges---skipping an FFN sub-layer is far more damaging than skipping an attention sub-layer (Table~\ref{tab:ASL-Skip})---so a comparison of raw values would assign almost every layer to attention and forfeit the per-layer discrimination visible in Table~\ref{tab:replaced_layers}, where five of the 32 decoder layers of LLaVA-1.5-7B (and seven of LLaVA-NeXT-7B) are designated FFN. Second, $m(\ell)$ is a per-layer property, whereas the sub-layer that a given token bypasses is fixed at the layer where that token enters skip mode and is then retained for all deeper layers. Tokens admitted at different layers therefore inherit different targets, so at any single layer some tokens may bypass attention while others bypass the FFN; this is what the bypass arcs in Fig.~\ref{overview} depict.

\subsection{Score Fusion and Cumulation}

The final skipping decision integrates both online and offline scores. At each layer $\ell$ (up to the network's midpoint, $L/2$), we compute a fused score:
\begin{equation}
    \mathrm{S}_{fuse}(i,\ell) = w_1 \cdot \mathbf{S}_{sa}(i,\ell) + w_2 \cdot \mathrm{SLC}^{Norm}(\ell),
\end{equation}
where $\mathrm{SLC}^{Norm}(\ell)\equiv\mathrm{SLC}^{Norm}_{m(\ell)}(\ell)$ is the score of that layer's designated skip target, and $w_2$ is fixed to $1$ throughout, so $w_1/w_2$ also sets the magnitude of $\mathrm{S}_{fuse}$. This score is accumulated as $\mathrm{S}_{skip}(i,\ell) = \mathrm{S}_{skip}(i,\ell-1) + \mathrm{S}_{fuse}(i,\ell)$. Once $\mathrm{S}_{skip}(i,\ell)$ exceeds a threshold $T_{\mathrm{skip}}$, token $i$ enters a ``skip mode'': its target is frozen as $\mathrm{target}(i)=m(\ell)$ and is then bypassed in every remaining layer up to $L$, while the accumulation itself is confined to $\ell \leq L/2$. Tokens not in skip mode proceed normally.

\section{Experiments}
\subsection{Experiment Settings}

\noindent\textbf{Models.} We evaluate our method on LLaVA-1.5-7B \cite{liu2024improved} and the high-resolution LLaVA-NeXT-7B \cite{liu2024LLaVAnext}. LLaVA-1.5 generates 576 visual tokens from 336×336 images. LLaVA-NeXT's sub-image partitioning strategy handles flexible resolutions, yielding 2,880 tokens for our evaluation, a 5$\times$ increase.

\noindent\textbf{Datasets.} We evaluate on various multimodal benchmarks, including GQA \cite{hudson2019gqa}, VQAv2 \cite{goyal2017making},
MME \cite{zhang2021mme}, TextVQA \cite{singh2019towards}, POPE \cite{li2023evaluating}, MMB \cite{liu2024mmbench} , MMVet \cite{yu2023mm}, MMStar \cite{chen2024we} and DocVQA \cite{mathew2021docvqa}. VQAv2 \cite{goyal2017making} evaluates visual recognition via open-ended questions on 265k MSCOCO images, with adversarially balanced answers to reduce language bias. We use the test-dev set (107k image-question pairs), scored against 10 human answers. GQA \cite{hudson2019gqa} tests structured reasoning using scene graph–generated questions on Visual Genome images. Evaluation is based on accuracy over 12.6k test-dev pairs. TextVQA \cite{singh2019towards} requires models to read and reason about OCR text in Open Images (e.g., signs, labels). Evaluated on 5k validation pairs. POPE \cite{li2023evaluating} measures object hallucination in MLLMs by querying object presence in MSCOCO images, reporting average F1 across three sampling strategies (8.9k pairs). MME \cite{zhang2021mme} assesses perception (OCR, object attributes, fine-grained recognition) via 14 binary subtasks; we report the perception score on 2,374 pairs. MMB \cite{liu2024mmbench} offers a hierarchical multimodal evaluation with multiple-choice questions across perception and reasoning. Both English (4,377) and Chinese (4,329) versions are used. MM-Vet \cite{yu2023mm} integrates six core capabilities (e.g., OCR, spatial reasoning, math) into 16 tasks, evaluated by ChatGPT on 218 diverse samples. MMStar \cite{chen2024we} provides a clean, human-curated benchmark of 1,500 visually dependent samples with minimal data leakage, measuring true multimodal gains across 6 capabilities and 18 axes. DocVQA \cite{mathew2021docvqa} focuses on document understanding, requiring fine-grained textual and spatial reasoning over 50k questions on 12k+ document images.

\noindent\textbf{Baselines.} We benchmark SPIDER against training-free efficient MLLM methods. Token pruning methods include FastV \cite{chen2024image}, which prunes a fixed ratio $R$ of visual tokens post-layer $K$ based on attention scores; VTW \cite{lin2025boosting}, which discards all visual tokens after layer $K$; and VisPruner \cite{zhang2024beyond}, which retains $V$ visual tokens from the vision encoder. Layer skipping method includes ShortV \cite{yuan2025shortv}, which replaces $l$ LLM layers with its ShortV layers. Our SPIDER retains $N$ visual tokens after the pruning stage and then allows $R_s$ of them to skip certain sub-layer modules. All the methods are compared at similar FLOPs reduction ratios. 

\begin{table*}
\centering
\caption{Comparison of training-free MLLM efficiency methods. FLOPs Ratio denotes the proportion of FLOPs retained relative to the vanilla model. Best results are in \textbf{bold}.}
\label{table1}
\resizebox{1.99\columnwidth}{!}{
\begin{tabular}{l |l l |c c c c c c c c c |c}
\toprule
\textbf{Method} & \textbf{TFLOPs} & \textbf{Ratio} & \textbf{VQAv2} & \textbf{GQA} & \textbf{MMStar} & \textbf{MME} & \textbf{MMB} & \textbf{POPE} & \textbf{MMVet} & \textbf{TextVQA} & \textbf{DocVQA} & \textbf{Acc. (\%)}\\
\midrule

\multicolumn{13}{c}{\textit{\textbf{LLaVA-1.5-7B (Upper Bound, All 576 Visual Tokens)}}} \\

\rowcolor{gray!25}
Vanilla & 8.5 & 100\% & 76.5 & 61.9  & 33.7 & 1510.7 & 64.1 & 85.9 &31.1  &58.2 & 21.5 & 100\%\\
\midrule
\textit{Approximately 55$\%$ TFLOPs}\\
\midrule
FastV ($K=2, R=50\%$) & 4.9 & 58\% & 73.5 & 60.2  & 32.4 & 1475.6 & 64.3 & 84.0 &29.8 & 57.2& 17.3 & 95.54\%\\
VTW ($K=16$) & 4.7 & 55\% & 66.3 & 55.1  & 32.8 & 1497.0 & 64.0 &  82.8& 19.2& 55.3& 16.2 & 88.93\%\\
ShortV ( $l=19$) & 4.7 & 55\% & 75.7 & \textbf{60.9}  & 33.3 & \textbf{1503.1} & 64.8 & 86.2 &27.9 &55.1 & 17.9 & 96.08\%\\
VisPruner ($V=288$) & 4.7 & 55\%   & 76.3 &60.8  & 33.3 &  1477.9 &63.7  & 86.3 &30.3 & 57.8 & 20.7 & 98.60\%\\

\rowcolor{blue!11}
SPIDER ($N=384, R_s=55\%$) &4.8  &  56\%  & \textbf{76.6} &\textbf{60.9}  & \textbf{34.9} &1498.5  &\textbf{64.9}  &\textbf{86.6} &\textbf{30.7} &\textbf{57.9}& \textbf{20.9} & \textbf{99.87\%}  \\  

\midrule
\textit{Approximately 25-30$\%$ TFLOPs}\\
\midrule
FastV ($K=2, R=75\%$) & 2.6 &  30\%  &74.3  & 56.6 & 30.8 & 1394 & 62.3 & 79.2 &30.3 &56.2 & 16.2 &92.33\%  \\
ShortV ( $l=31$) & 2.1 &  25\%  &56.1  & 47.7 & 29.3 & 771.5 & 56.1 & 58.5 & 17.2& 35.7 & 9.2 & 67.76\% \\
VisPruner ($V=128$) & 2.3 &  27\%  & 75.8 & 58.2 & 32.9 &\textbf{1461.4}  & 62.7 &84.6  &28.6 &57.0 & 18.0 & 95.27\% \\
\rowcolor{blue!21}
SPIDER ($N=144, R_s=35\%$) & 2.2 &  26\%  & \textbf{76.0} &\textbf{58.6}  & \textbf{33.7} &1458.2  & \textbf{63.7} & \textbf{84.8} & \textbf{30.3} & \textbf{57.3} & \textbf{18.5} & \textbf{96.73\%}  \\

\midrule
\multicolumn{13}{c}{\textit{\textbf{LLaVA-NeXT-7B (Upper Bound, All 2880 Visual Tokens)}}} \\

\rowcolor{gray!25}
Vanilla & 42.7 & 100\% & 80.0 & 62.9 & 37.1  & 1519.0 & 67.1 &86.3 &38.5 &59.6 & 68.4 &100\%\\
\midrule
\textit{Approximately 50$\%$ TFLOPs}\\
\midrule
FastV ($K=2, R=50\%$) & 22.0 & 52\% & 79.5 & 63.0 & 36.5  & 1482.0 & 66.3 & 86.5 &36.8 &58.1 & 59.0 & 97.10\%\\

VTW ($K=16$) & 21.8 & 51\% & 75.6 & 55.8 & 37.6 & 1518.2 & 67.1 & 84.9 & 18.5 &57.3 & 58.2 &89.71\% \\

ShortV ($l=19$) & 21.6 & 51\% & 78.8 & \textbf{63.4} & \textbf{37.8} & \textbf{1525.1} & \textbf{67.2} &86.9  &31.7 &58.3 & 59.8 &96.67\% \\

VisPruner ($V=1600$) & 21.8 &  51\%  & 79.9 & 62.5 & 37.3 & 1493.1 &66.7  & 88.0 & \textbf{37.3}&59.4 & 63.8 &98.81\%   \\
\rowcolor{yellow!21}
SPIDER ($N=1920, R_s=50\%$) & 21.9 & 51\%   &\textbf{80.2}  &62.6  &37.7  & 1510.5 &66.6  &\textbf{88.2} &36.6 &\textbf{59.7} & \textbf{64.6} & \textbf{99.11\%}\\
\midrule
\textit{Approximately 20-25$\%$ TFLOPs}\\
\midrule
FastV ($K=2, R=89\%$) & 8.5 & 20\%   & 71.9 &  55.9& 32.1 &1282.9 &53.4  & 71.7 &25.9 &55.7 & 43.7 &81.89\% \\

ShortV ( $l=29$) & 9.7 &  23\%  &58.6  &49.7  & 30.4 & 884.5 &51.2  &56.6  &21.5 &36.5 & 14.1 &63.56\%  \\
VisPruner ($V=640$) & 9.1 & 21\%   & \textbf{79.8} & 61.4 &36.5  &1490.8  & 65.2 & 85.9  &\textbf{36.7} &59.3& 50.6 &95.49\%  \\
\rowcolor{yellow!41}
SPIDER ($N=710, R_s=30\%$) &9.0  &  21\%  & \textbf{79.8} &\textbf{61.8}  & \textbf{37.3} & \textbf{1492.2} &\textbf{65.7}  & \textbf{87.8} &35.9 &\textbf{59.5}& \textbf{51.5} & \textbf{96.09\%}  \\

\bottomrule
\end{tabular}}
\end{table*}

\begin{table}
\caption{Comparisons of our MSV-Prune, with other SOTA training-free token pruning methods. Best results are in \textbf{bold}.}
\centering
\setlength{\tabcolsep}{4pt}
\renewcommand{\arraystretch}{1.2}
\resizebox{0.95\columnwidth}{!}{
\begin{tabular}{l|ccc|c}
\toprule
\textbf{Method} & \textbf{GQA} & \textbf{TextVQA} & \textbf{POPE} & \textbf{Acc. (\%)}  \\
\midrule
\rowcolor{gray!25}
\multicolumn{5}{c}{\emph{Upper Bound, All 2880 Tokens (100\%)}} \\
LLaVA-NeXT-7B  & 62.9 & 59.6 & 86.3   & 100.0\% \\
\midrule
\rowcolor{gray!25}
\multicolumn{5}{l}{\emph{Retain 320 Tokens} ( $ \downarrow $  88.9\%)} \\
FastV & 55.9 & 55.7 & 71.7  & 88.47\% \\
SparseVLM  & 56.5 & 52.4 & 73.5  & 87.64\% \\
VisionZip  & 58.1 & 57.6 & 75.0  & 91.97\% \\
VisPruner   & 58.4 & 57.6 & 80.4  & 94.22\% \\
MSV-Prune \textbf{(Ours)}  & \textbf{59.0} & \textbf{58.1} & \textbf{83.3} &   \textbf{95.93\%}  \\
\midrule
\rowcolor{gray!25}
\multicolumn{5}{l}{\emph{Retain 160 Tokens} ( $ \downarrow $  94.4\%)} \\
FastV  & 49.8 & 51.9 & 51.7  & 75.39\% \\
SparseVLM  & 50.2 & 45.1 & 54.6  & 72.92\% \\
VisionZip & 54.3 & 54.7 & 59.4  & 82.31\% \\
VisPruner   & 54.7 & 56.0 & 72.9  & 88.46\% \\
MSV-Prune \textbf{(Ours)}  & \textbf{55.1} & \textbf{57.1} & \textbf{73.3} & \textbf{89.45\%}   \\
\bottomrule
\end{tabular}}
\label{tab:MSV-Prune}
\end{table}

\begin{table}
\caption{Comparisons of our ASL-Skip with other training-free layer skipping/pruning methods. No visual token is pruned. Most of these methods are evaluated at around 80  $ \% $  of the original TFLOPs for fair comparisons. Best results are in \textbf{bold}. ``R" denotes TFLOPs ratio.}
\centering
\setlength{\tabcolsep}{4pt}
\renewcommand{\arraystretch}{1.2}
\resizebox{0.99\columnwidth}{!}{
\begin{tabular}{l|c|ccc|c}
\toprule
 \textbf{Method}  & \textbf{R}  & \textbf{MMStar} & \textbf{TextVQA} & \textbf{MME} & \textbf{Acc. (\%)}  \\
\midrule
\rowcolor{gray!25}
\multicolumn{6}{c}{\emph{Upper Bound, All 576 Tokens (100\%)}} \\
LLaVA-1.5-7B  & 100\% &  33.7 & 58.2 & 1510.7& 100\%\\
\midrule
ShortV   & 81\%& 33.8 & 57.3 &1503.3 & 99.42\%\\
Skip All Attn & 82\%& \textbf{34.1} & 51.1 &1300.6 & 91.69\%\\
Skip All FFN & 38\%&  28.9&  40.9& 875.9&  71.34\%\\
Skip Partial FFN & 81\%&  32.6&  57.4& 1483.1&  97.84\%\\
ASL-Skip \textbf{(Ours)}   &80\% & 33.7 & \textbf{58.1} &\textbf{1504.9}  & \textbf{99.81\%}\\
\bottomrule
\end{tabular}}
\label{tab:ASL-Skip}
\end{table}

To perform an ablation study on SPIDER's components, we introduce additional specialized baselines. We evaluate our token pruning module, MSV-Prune, against other token pruners like SparseVLM \cite{zhangsparsevlm} and VisionZip \cite{yang2025visionzip}. To assess the efficacy of our layer skipping module, ASL-Skip, we benchmark it against ShortV \cite{yuan2025shortv} and naive strategies that uniformly skip all attention or FFN modules.

\noindent\textbf{Aggregate Accuracy Metric.}
To summarize performance across benchmarks with different native scales, we report an aggregated relative accuracy metric, denoted as \textbf{Acc. (\%)}, defined as
\begin{equation}
\mathrm{Acc.\ (\%)}=
\frac{1}{|\mathcal{B}|}
\sum_{b\in\mathcal{B}}
\frac{s_b^{\mathrm{method}}}{s_b^{\mathrm{vanilla}}}\times 100,
\end{equation}
where $\mathcal{B}$ denotes the set of benchmarks included in the corresponding table, $s_b^{\mathrm{method}}$ is the score of the evaluated method on benchmark $b$, and $s_b^{\mathrm{vanilla}}$ is the score of the corresponding vanilla model on the same benchmark. This formulation first normalizes each benchmark by its vanilla performance and then averages across benchmarks, making the aggregate metric comparable despite different score ranges. For MME, we directly use the reported perception score as $s_b$ and normalize it by the vanilla MME perception score in the same manner as other benchmarks. All reported numbers are single-run results obtained under a fixed random seed with deterministic decoding; when comparing methods at matched compute, differences below roughly $0.3$ points should therefore be read as parity rather than as a strict ordering.

\noindent\textbf{Implementation Details for Specific Architectures.} 
For Qwen-series models with a PatchMerger, MSV-Prune extracts intermediate and final ViT hidden states and passes both through the same PatchMerger to obtain post-merger representations at two semantic depths. All importance computation and token selection are performed on these post-merger tokens, ensuring no conflict with the spatial merging structure. DeepStack feature tensors share the same token layout as the main stream by design, so the retained token indices apply to them directly. ASL-Skip operates entirely within the LLM decoder on the retained post-merger tokens and is independent of the vision encoder structure, so no additional adaptation is needed for Qwen-series models.



For LLaVA-NeXT with AnyRes, MSV-Prune is applied independently to each tile in the shared ViT feature space before the multi-modal projector. The base tile and all local tiles share the same pruning rule and token budget, with no extra quota reserved for the global tile. Per-tile masks are propagated together with the standard AnyRes reshape and unpadding operations, and the retained base-tile and local-tile tokens are then concatenated into the flat visual sequence consumed by the LLM decoder.

For Qwen3-VL-8B-Instruct, the number of visual tokens is input-dependent, so SPIDER is configured by the visual retention ratio $R_v$ rather than a fixed retained-token number. The two settings in Table~\ref{tab:qwen3-vl} use $R_v=0.5$ and $R_v=0.4$, respectively, and are chosen to approximately match the compared baselines in end-to-end FLOPs. ASL-Skip is token-adaptive rather than layer-fixed; empirically, most retained visual tokens begin skipping after the first four decoder layers, with an overall skipped-token proportion of around $70\%$ and $60\%$, respectively. FLOPs are measured over the full multimodal pipeline under the same prompt and image preprocessing setup.

\subsection{Quantitative Results}
We apply SPIDER to the classic LLaVA-1.5 and LLaVA-NeXT models and comprehensively compare against prior approaches.
As shown in Table~\ref{table1}, SPIDER consistently matches or surpasses baselines across multiple benchmarks at similar or lower FLOPs, outperforming other training‑free methods and achieving the best trade‑off between efficiency and performance. Notably, even under a relatively aggressive FLOPs reduction ratio to $20\%$, SPIDER applied on LLaVA-NeXT-7B retains $96.09\%$ of the overall performance, whereas skip-layer-based baselines such as ShortV degrade substantially. On challenging benchmarks such as MMVet and MMStar, which demand strong spatial awareness and reasoning capabilities, SPIDER maintains competitive accuracy, underscoring its robustness under high compression.

More specifically, we ablate the two core components of SPIDER separately: MSV-Prune in Table~\ref{tab:MSV-Prune} and ASL-Skip in Table~\ref{tab:ASL-Skip}.

Table~\ref{tab:MSV-Prune} shows that MSV-Prune consistently surpasses SOTA pruners across all retention ratios. This confirms that relying solely on the vision encoder's last layer introduces a semantic focus shift where more fine-grained object-centric information is sacrificed for abstract global context. By incorporating middle-layer features as semantic anchors, MSV-Prune ensures a more holistic representation of the visual scene, which is reflected in the significantly higher scores on OCR-centric tasks like TextVQA.

Table~\ref{tab:ASL-Skip} reveals that ASL-Skip preserves overall performance more effectively than existing layer-pruning methods at the same computational cost, underscoring the importance of fine-grained sub-layer skipping. Even under identical FLOPs ratios, non-adaptive partial FFN skipping still incurs a 2\% accuracy drop compared to ASL-Skip, while skipping all FFN sub-layers leads to nearly 30\% degradation. These results indicate that skipping attention sub-layers is significantly less harmful than skipping FFNs, and that adaptive skipping (ASL-Skip) is the optimal strategy.



To further demonstrate the architecture-agnostic nature of SPIDER, we extend our 
evaluation to Qwen2.5-VL-3B-Instruct and Qwen3-VL-8B-Instruct. As shown in 
Table~\ref{tab:qwen} and Table~\ref{tab:qwen3-vl}, SPIDER maintains competitive overall accuracy across different TFLOPs ratios on both models. On Qwen3-VL-8B-Instruct, 
we further compare against more recent methods including PruneSID~\cite{fang2026prune}, 
IVC-Prune~\cite{sun2026ivc}, iLLaVA~\cite{hu2024illava}, and ERASE~\cite{lee2026erase}. 
SPIDER retains 99.06\% and 98.33\% of the vanilla performance at the approximately 
55\% and 45\% TFLOPs settings, respectively. At matched compute, SPIDER is on par 
with the strongest recent baseline in aggregate accuracy, while leading on ChartQA 
and DocVQA under both settings and also obtaining the best GQA score at the 45\% 
TFLOPs setting. These results confirm that the redundancy patterns targeted by 
SPIDER, namely the semantic focus shift across vision encoder layers and the sub-layer heterogeneity within the LLM decoder, are intrinsic properties of the MLLM paradigm that generalize across model families rather than being artifacts of specific architectures.

Beyond its training-free mode, SPIDER adapts to a training-aware framework for performance enhancement. Table~\ref{tab:train_mode} shows results after fine-tuning on LLaVA-1.5-7B using LoRA for 1 epoch with 665K instruction data, following the same training setting as LLaVA-PruMerge \cite{shang2024LLaVA}. Notably, the training-free SPIDER already matches PruMerge+, which requires fine-tuning, under the aggregate metric of Eq.~(7) (both $99.0\%$), leading on POPE ($84.8$ vs.\ $84.0$) and TextVQA ($57.4$ vs.\ $57.1$) while trailing on MMB ($63.9$ vs.\ $64.9$); post-training refinement then elevates SPIDER to $99.5\%$.

\subsection{Qualitative Results}
\begin{figure*}
    \centering \includegraphics[width=0.99\textwidth,height=0.42\textwidth]{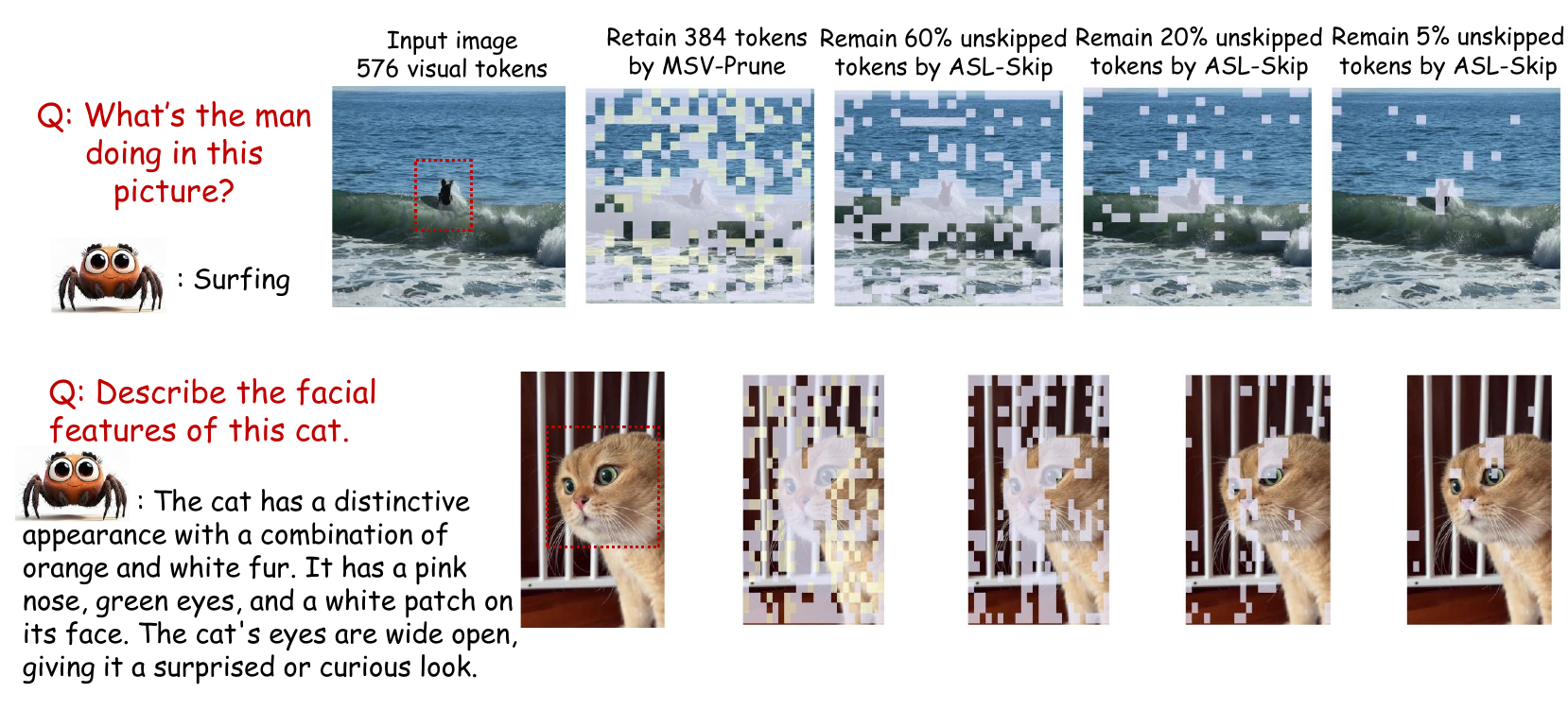}
    \caption{Visualization of retained tokens by MSV-Prune and unskipped tokens by ASL-Skip; anchors in purple ($r=0.7$), complementary tokens in yellow.
}\label{vis}
\end{figure*}

Figure~\ref{vis} shows MSV‑Prune retains semantically critical tokens, while ASL‑Skip progressively skips less informative ones. Even at aggressive skip rates, core semantics are maintained, enabling accurate scene and object queries. 

\begin{figure}
    \centering \includegraphics[width=0.49\textwidth,height=0.32\textwidth]{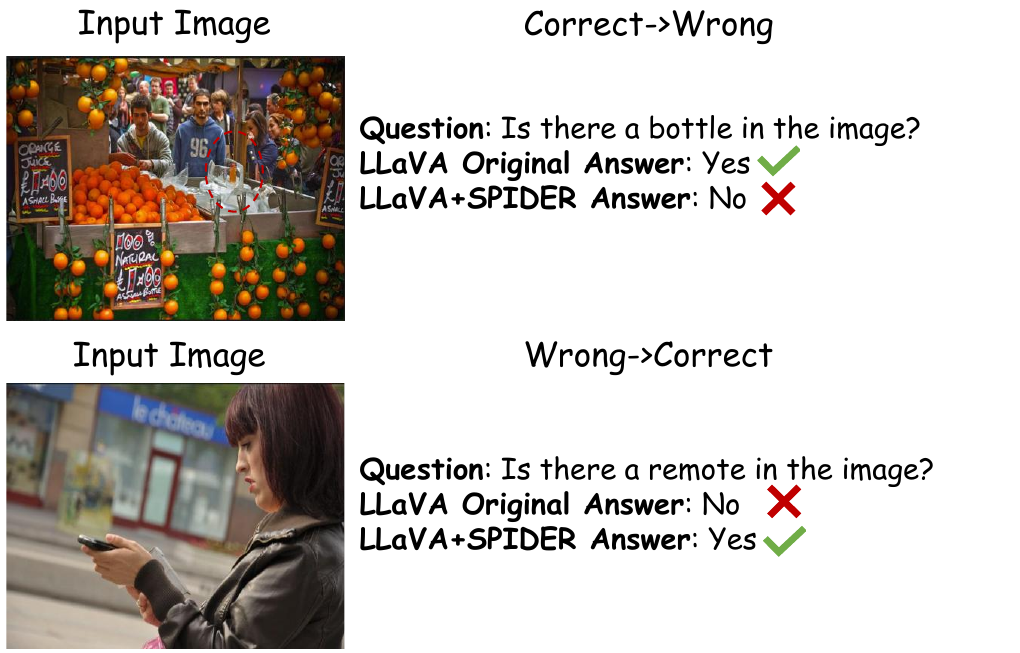}
    \caption{Instance-level visualization of boundary shift.
}\label{boundary_shift}
\end{figure}

We additionally explore the extent of knowledge boundary drift caused by SPIDER via conducting a fine-grained instance-level analysis on the POPE test set using LLaVA-1.5-7B with our SPIDER method (at 56$\%$ of original FLOPs). Accuracy improves from 85.9$\%$ to 86.6$\%$, reflecting a net gain: the number of originally incorrect predictions corrected by SPIDER exceeds the number of originally correct ones flipped to wrong by 0.7$\%$. Qualitative examples are provided in Fig.~\ref{boundary_shift}. This gain stems from SPIDER’s retention of mid-layer tokens encoding object-centric regions, which enhances discriminative grounding and reduces false positives—particularly in hallucination-sensitive queries.

\begin{figure}
    \centering \includegraphics[width=0.49\textwidth,height=0.16\textwidth]{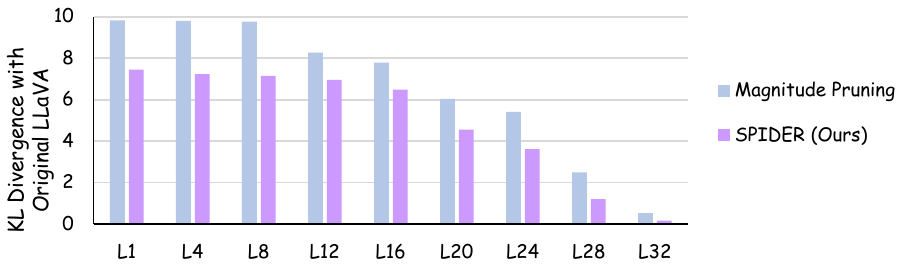}
    \caption{KL divergence comparison with magnitude pruning across LLM layers.
}\label{compare_magprune}
\end{figure}

\begin{figure}
    \centering \includegraphics[width=0.49\textwidth,height=0.13\textwidth]{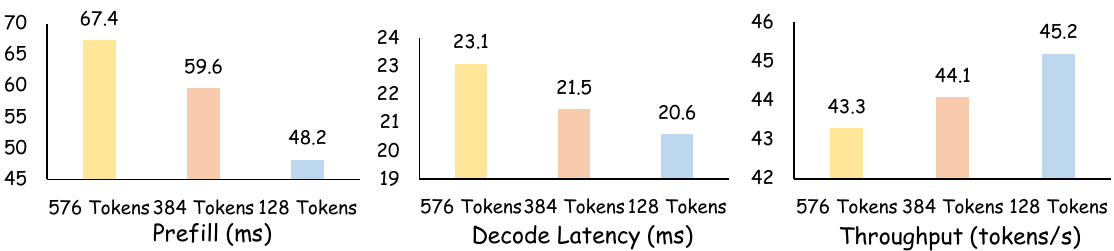}
    \caption{SPIDER efficiency on LLaVA-1.5-7B at varying token reduction ratios; key image regions relevant to queries marked by red dashed boxes.
}\label{efficiency}
\end{figure}

\subsection{Efficiency Analysis}
Figure~\ref{efficiency} demonstrates the efficiency gains of SPIDER across varying token reduction ratios on LLaVA-1.5-7B. As the number of retained visual tokens decreases from 576 to 128, both prefill and decode latencies are significantly reduced---prefill latency drops from 67.4~ms to 48.2~ms (a 28.5\% reduction), and decode latency decreases from 23.1~ms to 20.6~ms (a 10.8\% reduction). Concurrently, throughput increases from 43.3 to 45.2 tokens per second, indicating improved inference speed under lower computational load. These results highlight SPIDER's ability to achieve substantial latency reduction and higher throughput while maintaining strong semantic fidelity, as evidenced by its consistent performance in downstream tasks even at aggressive token pruning rates. This balance between efficiency and accuracy makes SPIDER suitable for real-time vision-language applications where low-latency responses are essential.

The larger gain in prefill relative to decoding is expected: prefill computation scales directly with sequence length and is dominated by parallel matrix operations, whereas autoregressive decoding processes one token at a time and is increasingly bottlenecked by KV-cache access and memory bandwidth rather than arithmetic throughput. For ASL-Skip, the token-level dynamic sparsity introduced by our method is irregular and input-dependent, which standard dense GPU kernels cannot fully exploit without hardware-aware sparse implementations. We note this as a current practical limitation and discuss token regrouping and sparse kernel support as directions for future decode-stage acceleration.

\begin{table}
\caption{Results on Qwen2.5-VL-3B-Instruct. Best results are in \textbf{bold}.}
\centering
\setlength{\tabcolsep}{4pt}
\renewcommand{\arraystretch}{1.2}
\resizebox{0.99\columnwidth}{!}{
\begin{tabular}{l| c c c c c |c}
\toprule
\textbf{Method} & \textbf{MMB} & \textbf{MMB}\textsuperscript{CN} & \textbf{POPE} & \textbf{SQA}\textsuperscript{IMG} & \textbf{VizWiz} & \textbf{Acc. (\%)} \\
\midrule
\rowcolor{gray!25}
\multicolumn{7}{c}{\emph{Vanilla, 100\% Tokens}} \\
Qwen2.5-VL-3B-Instruct & 77.3 & 73.0 & 87.0 & 80.4 & 68.3 & 100 \% \\
\midrule
\multicolumn{7}{l}{\emph{Approximately 35 $ \% $  TFLOPs}} \\
\midrule
FastV & 74.4 & 70.6 & 85.0 & 79.3 & 66.9 & 97.4\% \\
VisionZip & 74.9 & 69.8 & 85.4 & 80.1 & 67.1 & 97.7\% \\
HiPrune & 75.8 & 71.3 & 86.0 & 80.0 & 67.5 & 98.6\% \\
SPIDER \textbf{(Ours)}  &\textbf{76.2} & \textbf{72.0} & \textbf{86.3} & \textbf{80.1} &\textbf{67.8}  & \textbf{99.1\%} \\
\midrule
\multicolumn{7}{l}{\emph{Approximately 25  $ \% $  TFLOPs}} \\
\midrule
FastV & 72.4 & 69.2 & 82.7 & 79.6 & 66.2 & 95.9\% \\
VisionZip & 73.5 & 67.4 & 84.6 & 80.0 & 66.3 & 96.2\% \\
HiPrune & 74.0 & 69.3 & 84.7 & \textbf{80.3} & 66.5 & 97.1\% \\
SPIDER \textbf{(Ours)}  &\textbf{75.3} & \textbf{69.8} &\textbf{85.2}  & 79.8 & \textbf{67.4} &  \textbf{97.8\%}\\
\bottomrule
\end{tabular}}
\label{tab:qwen}
\end{table}

\subsection{Comparison with Magnitude Pruning}
Prior work has shown that token pruning can induce knowledge boundary drift, which means instances once answered correctly may become incorrect after pruning, and vice versa, posing risks in applications requiring reliable responses to critical queries. To address this, we go beyond conventional benchmarks and perform an instance-level analysis of SPIDER's semantic fidelity on the POPE test set using LLaVA-1.5-7B at 56\% of original FLOPs. Accuracy improves from 85.9\% to 86.6\%, a net gain of 0.7\%, as SPIDER corrects more originally wrong predictions than it flips correct ones. This stems from its retention of mid-layer tokens encoding object-centric regions, enhancing discriminative grounding and reducing false positives, especially in hallucination-sensitive queries. We also compare SPIDER with magnitude pruning: for each layer $\ell$ and visual token $i$, we compute L1 norms of attention and FFN outputs, average them into layer-wise scores $M_{\text{Attn}}(\ell)$ and $M_{\text{FFN}}(\ell)$, and prune the lower-magnitude sub-module to obtain a magnitude-pruned LLaVA. KL divergence between per-layer logits of the pruned and original models quantifies post-pruning distortion. As shown in Figure~\ref{compare_magprune}, SPIDER achieves lower KL divergence, confirming its superior layer-skipping strategy with minimal degradation.

\begin{table}
\caption{Results on Qwen3-VL-8B-Instruct. Best results are in \textbf{bold}.}
\centering
\setlength{\tabcolsep}{4pt}
\renewcommand{\arraystretch}{1.2}
\resizebox{0.99\columnwidth}{!}{
\begin{tabular}{l| c c c c|c}
\toprule
\textbf{Method} & \textbf{TextVQA} & \textbf{ChartQA} & \textbf{DocVQA} & \textbf{GQA} & \textbf{Acc. (\%)} \\
\midrule
\rowcolor{gray!25}
\multicolumn{6}{c}{\emph{Vanilla, 100\% Tokens}} \\
Qwen3-VL-8B-Instruct & 82.98 & 83.16 & 95.75 & 61.88 & 100\% \\
\midrule
\multicolumn{6}{l}{\emph{Approximately 55\% TFLOPs}} \\
\midrule
PruneSID & 73.46 & 63.56 & 90.51 & 60.08 & 89.15\% \\
IVC-Prune & \textbf{82.35} & 79.60 & 95.31 & 61.33 & 98.40\% \\
iLLaVA & 76.11 & 64.68 & 84.89 & 61.15 & 89.25\% \\
ERASE & 82.12 & 81.44 & 95.44 & \textbf{61.37} & 98.94\% \\
SPIDER \textbf{(Ours)} & 82.27 & \textbf{81.68} & \textbf{95.50} & 61.35 & \textbf{99.06\%} \\
\midrule
\multicolumn{6}{l}{\emph{Approximately 45\% TFLOPs}} \\
\midrule
PruneSID & 69.98 & 57.32 & 85.77 & 60.06 & 84.97\% \\
IVC-Prune & 81.74 & 76.48 & 94.97 & 61.11 & 97.11\% \\
iLLaVA & 72.57 & 61.72 & 79.63 & 60.91 & 85.81\% \\
ERASE & 81.77 & 79.96 & 95.00 & 61.31 & 98.25\% \\
SPIDER \textbf{(Ours)} & \textbf{81.82} & \textbf{80.03} & \textbf{95.16} & \textbf{61.33} & \textbf{98.33\%} \\
\bottomrule
\end{tabular}}
\label{tab:qwen3-vl}
\end{table}

\begin{table}
\caption{Comparisons of training-aware modes. The TFLOPs are kept at a similar level around 30 $ \% $  for fair comparison, where 1/4 of visual tokens are preserved for PruMerge+. Best results are in \textbf{bold}.}
\centering
\setlength{\tabcolsep}{4pt}
\renewcommand{\arraystretch}{1.2}
\resizebox{0.99\columnwidth}{!}{
\begin{tabular}{l|ccc|c}
\toprule
 \textbf{Method}    & \textbf{POPE} & \textbf{TextVQA} & \textbf{MMB} & \textbf{Acc. (\%)}  \\
\midrule
\rowcolor{gray!25}
\multicolumn{5}{c}{\emph{Upper Bound, All 576 Tokens (100\%)}} \\
LLaVA-1.5-7B  &  85.9  & 58.2 &64.1 & 100\%\\
\midrule
\multicolumn{5}{l}{\emph{Approximately 30 $ \% $  TFLOPs}} \\
\midrule
+ PruMerge+ (Train)   & 84.0 & 57.1 & \textbf{64.9} & 99.0\%\\
+ SPIDER (Train-Free)   & 84.8 & 57.4 & 63.9 & 99.0\%\\
+ SPIDER (Train)   & \textbf{85.0}  & \textbf{57.6} &64.4  & \textbf{99.5\%}\\
\bottomrule
\end{tabular}}
\label{tab:train_mode}
\end{table}

\subsection{Ablation study and analysis}

\begin{table}
\caption{Ablation of our token pruning method, MSV-Prune, on LLaVA-NeXT-7B. Best results are in \textbf{bold}. `*' denotes our default setting.}
\centering
\setlength{\tabcolsep}{4pt}
\renewcommand{\arraystretch}{1.2}
\resizebox{0.99\columnwidth}{!}{
\begin{tabular}{l|ccc|c}
\toprule
\textbf{Method} & \textbf{GQA} & \textbf{TextVQA} & \textbf{POPE} & \textbf{Acc. (\%)}  \\
\midrule
\rowcolor{gray!25}
\multicolumn{5}{c}{\emph{Upper Bound, All 2880 Tokens (100\%)}} \\
LLaVA-NeXT-7B  & 62.9 & 59.6 & 86.3   & 100.0\% \\
\midrule
\rowcolor{gray!25}
\multicolumn{5}{c}{\emph{Retain 1920 Tokens}} \\
\midrule
\rowcolor{purple!15}
\multicolumn{5}{c}{\emph{(a) Semantic Cluster}} \\
w/o Semantic Cluster & 61.3 & 59.3 &  87.2 & 99.33\% \\
Cluster Num =2 &61.9  & 59.4 & 87.9  & 99.98\% \\
Cluster Num =8 & 61.6 & 59.3 &  87.7 & 99.68\% \\
Cluster Num =4* & \textbf{62.6} & \textbf{59.7} &  \textbf{88.2} & \textbf{100.64\%} \\
\midrule
\rowcolor{purple!15}
\multicolumn{5}{c}{\emph{(b) Multi-Layer Tokens}} \\
w/o Middle Layer Tokens  & 62.2 & 58.5 &  86.6 & 99.13\% \\
only Middle Layer Tokens  & 61.9 & 58.7 & 87.1 & 99.28\% \\
w/ Middle Layer Tokens*   & \textbf{62.6} & \textbf{59.7} &  \textbf{88.2} & \textbf{100.64\%} \\
\midrule
\rowcolor{purple!15}
\multicolumn{5}{c}{\emph{(c) Components of  $ S_{ij} $ }} \\
w/o Similarity with  $ T_v^{anc} $   & 62.5 & 59.3 & 87.8  & 100.20\% \\
w/o Inter-Layer Similarity   & 61.9 & 59.0& 87.5& 99.60\%\\
cos  $ \rightarrow $  MSE & 62.3 & 59.2 & 87.7 & 99.99\% \\
cos  $ \rightarrow $  KL & 62.2 & 59.1 & 87.4 & 99.77\% \\
All Equipped*   & \textbf{62.6} & \textbf{59.7} &  \textbf{88.2} & \textbf{100.64\%} \\
\bottomrule
\end{tabular}}
\label{ab:MSV-Prune}
\end{table}

\begin{table}
\caption{Ablation of our sub-layer skipping method, ASL-Skip. No visual token is pruned. Best results are in \textbf{bold}.}
\centering
\setlength{\tabcolsep}{4pt}
\renewcommand{\arraystretch}{1.2}
\resizebox{0.99\columnwidth}{!}{
\begin{tabular}{l|ccc|c}
\toprule
\textbf{Method} & \textbf{MMStar} & \textbf{TextVQA} & \textbf{MME} & \textbf{Acc. (\%)}  \\
\midrule
\rowcolor{gray!25}
\multicolumn{5}{c}{\emph{Upper Bound, All 576 Tokens (100\%)}} \\
LLaVA-1.5-7B  &  33.7 & 58.2 & 1510.7& 100\%\\
\midrule
\rowcolor{pink!30}
\multicolumn{5}{c}{\emph{(a) Components of  $ \mathbf{S}_{fuse} $ }} \\
w/o  $ \mathbf{S}_{sa} $  & 33.5 & 57.6 & 1497.8  & 99.17\% \\
w/o  $ \mathbf{SLC} $  & 32.7 & 58.0 & 1483.6  & 98.30\% \\
All Equipped* & \textbf{33.7}  & \textbf{58.1} & \textbf{1504.9} & \textbf{99.81\%}\\ 
\midrule
\rowcolor{pink!30}
\multicolumn{5}{c}{\emph{(b) Accumulation Mode of  $ S_{skip} $  }} \\
w/o Accumulation &33.4 & 57.8 & 1499.1    & 99.22\% \\
w/ Accumulation*   & \textbf{33.7}  & \textbf{58.1} & \textbf{1504.9} & \textbf{99.81\%}\\ 
\midrule
\rowcolor{pink!30}
\multicolumn{5}{c}{\emph{(c) Components of  $ \mathbf{S}_{sa} $ }} \\
w/o  $ \mathbf{E}_{ii} $  & 33.6 &  57.9&  1501.0 & 99.52\% \\
w/o  $ \mathbf{F}_{itc} $    &  33.5& 57.7& 1498.5&99.25\% \\
 $ \mathbf{S}_{sa} $ *   & \textbf{33.7}  & \textbf{58.1} & \textbf{1504.9} & \textbf{99.81\%}\\ 
\bottomrule
\end{tabular}}
\label{ab:ASL-Skip}
\end{table}

\begin{table}
\caption{Ablation of anchor token ratio in our token pruning method, MSV-Prune. Best results are in \textbf{bold}. '*' denotes our default setting.}
\centering
\setlength{\tabcolsep}{4pt}
\renewcommand{\arraystretch}{1.2}
\resizebox{0.95\columnwidth}{!}{
\begin{tabular}{c|ccc|c}
\toprule
\textbf{Anchor Ratio} & \textbf{GQA} & \textbf{TextVQA} & \textbf{POPE} & \textbf{Acc. (\%)}  \\
\midrule
LLaVA-1.5-7B  & 61.9 & 58.2 & 85.9   & 100.0\% \\
\midrule
0 & 59.3 & 54.7 & 83.7  & 95.74\% \\
0.3  & 60.2	& 57.0 &  86.2 & 98.51\% \\
0.5  & 60.6	 & 57.5 & 86.4  & 99.09\% \\
0.7*  & \textbf{60.9} & \textbf{57.9} & \textbf{86.6}  & \textbf{99.56\%} \\
1.0  & 60.8	 & 56.9 & 86.1 & 98.74\% \\
\bottomrule
\end{tabular}}
\label{ab:anchor}
\end{table}

\noindent\textbf{Ablation on MSV-Prune (Table~\ref{ab:MSV-Prune}).} Under a fixed token budget (1920 tokens, \( r=0.7 \)), removing any component of MSV-Prune (semantic clustering, middle-layer tokens, or similarity scores \( S_{ij} \)) reduces accuracy, highlighting the importance of modeling semantic shifts across layers. Notably, discarding middle-layer tokens hurts fine-grained tasks like OCR (e.g., TextVQA) more severely, and using only middle-layer tokens to compute intra-cluster similarity helps extracting object-centric cues. Replacing cosine similarity in \( S_{ij} \) with MSE or KL divergence also degrades performance: MSE is sensitive to non-semantic magnitude differences~\cite{bengio2013representation}, and KL divergence may not be the most appropriate choice here, as the layer features do not naturally form valid probability distributions. The full model (``All Equipped'') achieves the best average accuracy.

\noindent\textbf{Ablation on Anchor Token Ratio in MSV-Prune (Table~\ref{ab:anchor}).}
We perform a fine-grained ablation on the anchor token ratio to characterize its influence on model performance. The results reveal a clear non-monotonic trend: completely disabling anchor tokens (ratio = 0) incurs substantial drops in accuracy across all benchmarks (e.g., $-2.6$ on GQA, $-3.5$ on TextVQA), confirming that unstructured token removal disrupts critical visual grounding. In contrast, over-retaining anchors (ratio = 1.0) also degrades performance---particularly on TextVQA (56.9 vs. 57.9 at ratio = 0.7), suggesting that excessive token preservation hinders the model's ability to exploit sparsity and introduces redundant computation without meaningful gains. The optimal trade-off is achieved at an anchor ratio of 0.7, which not only attains peak scores on GQA (60.9), TextVQA (57.9), and POPE (86.6), but also maintains 99.56\% of the original model's effective capacity. This indicates that a moderate yet principled selection of anchor tokens enables MSV-Prune to preserve cross-modal alignment while maximizing computational savings. 

\begin{figure}
\centering
\includegraphics[width=0.99\columnwidth]{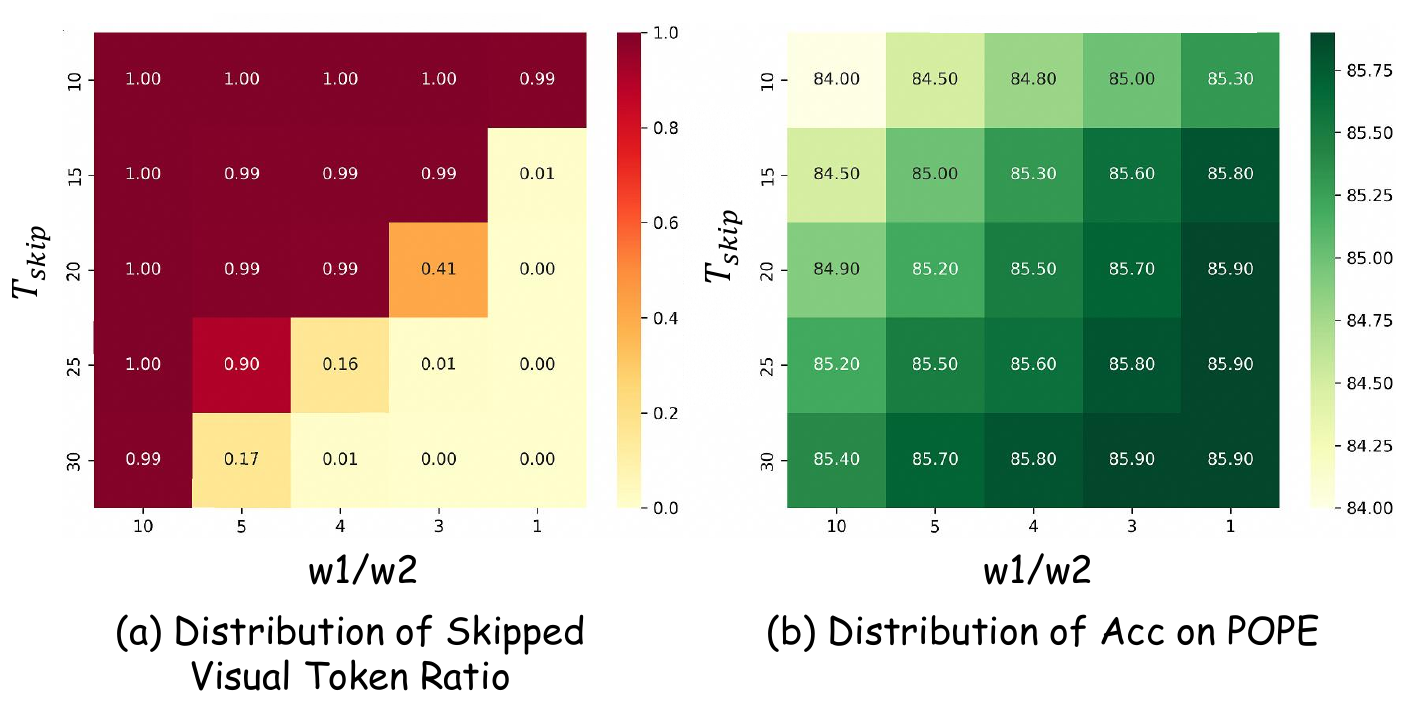}
\caption{Heatmap of token skipping proportion (a) and POPE accuracy (b) for different  $ T_{skip} $  and  $ w_1/w_2 $  settings on LLaVA-1.5-7B. No visual token is pruned.}
\label{ab:ratio}
\end{figure}

\noindent\textbf{Ablation on ASL-Skip (Table~\ref{ab:ASL-Skip}).} When ablating $\mathbf{S}_{fuse}$ in (a), removing the $\mathbf{S}_{sa}$ component forces all visual tokens to follow the offline SLC list, while omitting $\mathbf{SLC}$ makes skipped tokens bypass entire subsequent layers, leading to higher performance variance and reduced robustness. In (b) , the addition-based accumulation outperforms non-accumulative decisions, as tokens consistently deemed unimportant across consecutive layers are more reliably skip candidates than those judged by a single layer in isolation. In (c), both semantic uncertainty and image-text correlation are necessary for deciding if and when a token should start sub-layer skipping.

\noindent\textbf{Ablation on the Balance Between Token Skipping Ratio and Accuracy in ASL-Skip (Figure~\ref{ab:ratio}).} 
We evaluate how the cumulative skip threshold  $ T_{\text{skip}} $  and weight ratio  $ w_1/w_2 $  govern the balance between computational savings and output fidelity. Here,  $ T_{\text{skip}} $  sets the activation threshold for skipping (higher values suppress skipping), while  $ w_1/w_2 $  balances the online skippability score  $ S_{sa} $  (derived from token entropy  $ \mathbf{E}_{ii} $  and image-text correlation  $ \mathbf{F}_{itc} $ ) and the offline normalized sub-layer redundancy score  $ \mathrm{SLC}^{\text{Norm}} $ .


Figure~\ref{ab:ratio} (a) confirms an inverse relationship between  $ T_{\text{skip}} $  and skipping ratio: lowering  $ T_{\text{skip}} $  (e.g.,  $ T_{\text{skip}}=10 $ ) drastically increases skipping rates, with the entire  $ T_{\text{skip}}=10 $  row skipping essentially all tokens ($0.99$--$1.00$). With $w_2$ fixed, the weight ratio sets the scale of $\mathrm{S}_{fuse}$ in Eq.~(6), so a larger $ w_1/w_2 $  makes the accumulator cross $ T_{\text{skip}}$ earlier: at  $ w_1/w_2=10 $  the skipped ratio stays at $0.99$--$1.00$ for every  $ T_{\text{skip}} $ , whereas the small $ w_1/w_2 $  columns are the ones most sensitive to  $ T_{\text{skip}} $ . Conversely, raising  $ T_{\text{skip}} $  (e.g.,  $ T_{\text{skip}}=30 $ ) suppresses skipping for  $ w_1/w_2 \leq 5 $ , preserving nearly all tokens but yielding minimal acceleration, while the  $ w_1/w_2=10 $  column remains saturated.


POPE accuracy (Figure~\ref{ab:ratio} (b)) directly reflects this trade-off: the most aggressive configurations lie in the  $ T_{\text{skip}}=10 $  row, whose accuracies are the lowest in the grid and bottom out at $84.00\%$ at  $ w_1/w_2=10 $ , i.e.\ $1.9$ points below the dense baseline, due to excessive skipping of discriminative features. Overly conservative settings ( $ T_{\text{skip}}=30 $ ) maintain high accuracy but sacrifice efficiency gains. Reading the two panels jointly identifies the operating point: nine configurations lie within $0.2$ points of the dense LLaVA-1.5-7B POPE baseline of $85.9\%$, and among them ( $ T_{\text{skip}}=20 $ ,  $ w_1/w_2=3 $ ) attains $85.70\%$ with by far the highest skipped-token ratio, $0.41$; the runner-up ( $ T_{\text{skip}}=30 $ ,  $ w_1/w_2=5 $ ) reaches the same $85.70\%$ but skips only $0.17$. The four configurations that do reach $85.90\%$ all have a skipped-token ratio of $0.00$ in panel (a) and therefore accelerate nothing. We thus adopt ( $ T_{\text{skip}}=20 $ ,  $ w_1/w_2=3 $ ), which costs $0.2$ points on POPE while placing $41\%$ of the visual tokens in skip mode. This configuration leverages  $ S_{sa} $ 's context-aware assessment to retain semantically critical tokens and  $ \mathrm{SLC}^{\text{Norm}} $ 's precomputed redundancy profile to skip non-essential sub-layer computations.

These results establish two principles: (1)  $ T_{\text{skip}} $  must be calibrated to avoid under-skipping (wasted efficiency) or over-skipping (accuracy loss); (2) prioritizing the online score ( $ w_1/w_2 > 1 $ ) ensures skipping decisions adapt to input-specific semantics, while  $ w_1/w_2 $  must remain bounded, since at  $ w_1/w_2=10 $  the accumulator saturates and skipping becomes indiscriminate. 


\begin{figure*}
\centering
\includegraphics[width=1.99\columnwidth]{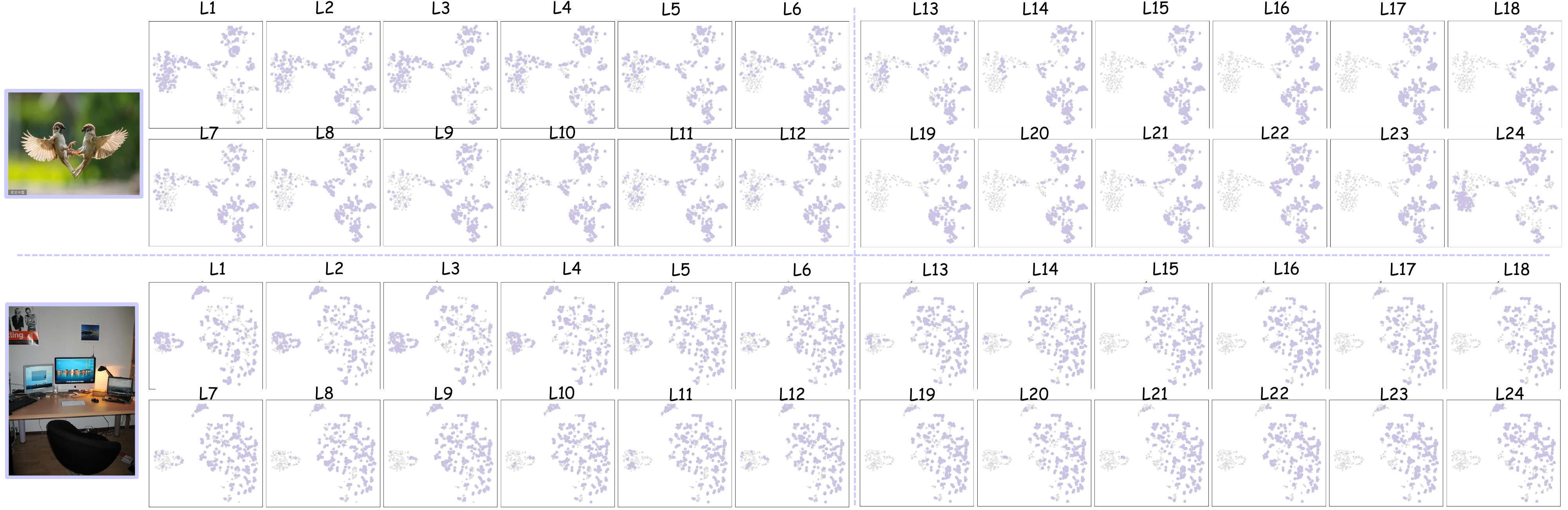}
\caption{Distribution of High-attention Tokens Across Vision Encoder Layers.}
\label{cluster}
\end{figure*}

\subsection{Distribution of High-attention Visual Tokens}
The visualization of top 50$\%$ high-attention tokens across all 24 layers of the CLIP vision encoder, as displayed in Figure~\ref{cluster}, reveals a clear progression in how visual information is processed and represented hierarchically. This hierarchical transformation provides a principled basis for our middle and deep layer selection in the vision encoder: we select the $12^{th}$ layer (middle layer) and the $24^{th}$ layer (final layer) as key feature extraction points.

In the early layers (Layers 1-6), high-attention tokens are evenly distributed across the embedding space, with no strong clustering patterns. This reflects the vision encoder’s initial focus on low-level features, such as edges, textures, and simple shapes—captured at a fine-grained level. These representations are highly local and lack semantic coherence. While rich in detail, they contain too much noise and redundancy for direct use in language modeling, where global semantics and structured concepts are essential.

At Layer 12, we observe a transitional state: the high-attention tokens begin to form distinct clusters while still maintaining some diversity. This indicates that the model has started combining low-level features into higher-level structures, such as object parts or spatial configurations, but without fully collapsing into coarse semantic categories. This makes it an ideal source for capturing the fine-grained object-centric details necessary for more information-dense tasks.

By Layer 24, the final layer before the output, high-attention tokens exhibit strong clustering, indicating that the vision encoder has synthesized the input into coherent semantic units corresponding to major objects or scene components. The representation becomes highly abstract and globally consistent, emphasizing holistic understanding rather than fine details.

\subsection{Replaced Sub-Layers}
For the selection of replaced layers, the $\mathbf{SLC}$ metric is computed based on a randomly sampled dataset comprising 150 cases, with 50 instances sampled equally from GQA, MMVet, and POPE. Layers are then replaced, either with VSkip-Attn or VSkip-FFN modules, in order of decreasing $\mathrm{SLC}^{Norm}$ of the designated sub-layer, i.e.\ the most skippable sub-layer first. 
Table~\ref{tab:replaced_layers} enumerates the layer IDs corresponding to the replaced components within the default SPIDER architecture.

\begin{table}
\caption{Replaced layers for different MLLM series and parameter scales.}
\centering
\resizebox{0.49\textwidth}{!}{%
\begin{tabular}{l p{0.6\linewidth}} 
\toprule
\textbf{Model Series} & \textbf{Replaced Sub-Layers} \\
\midrule
LLaVA-1.5-7B
& \textbf{Attention}: 25,27,28,30,23,26,22,24,21,0,20,3,\newline
18,4,19,17,14,15,5,16,12,1,13,7,8,9,10; \newline
\textbf{FFN}: 31,29,2,11,6 \\
\midrule
LLaVA-NeXT-7B
& \textbf{Attention}: 28,29,27,30,23,24,25,22,21,26,20,19,\newline 
18,17,15,16,14,12,4,13,5,0,7,6,8; \newline
\textbf{FFN}: 31,1,2,3,11,9,10 \\
\bottomrule
\end{tabular}
}
\label{tab:replaced_layers}
\end{table}

\subsection{Sensitivity Analysis on Middle Layer Selection}
\begin{figure}
    \centering
    \includegraphics[width=0.49\textwidth]{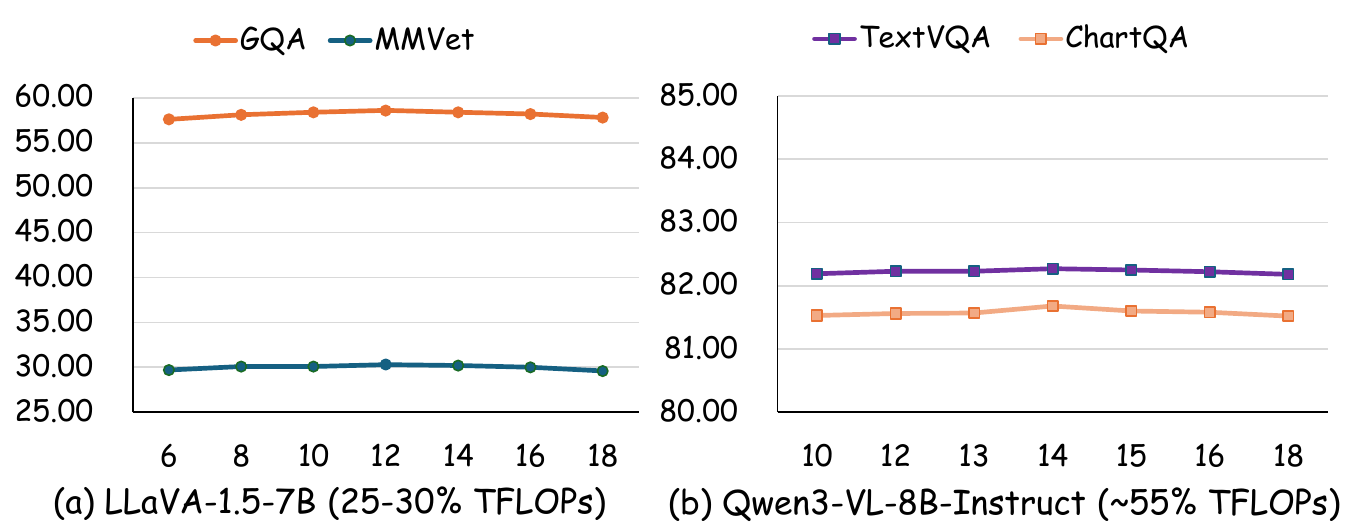}
    \caption{Sensitivity of MSV-Prune to the choice of intermediate vision encoder layer on (a) LLaVA-1.5-7B (25--30$\%$ TFLOPs) and (b) Qwen3-VL-8B-Instruct ($\sim$55\% TFLOPs). Performance remains stable across a wide range of layer choices on both architectures, confirming that the midpoint layer $\lfloor L/2 \rfloor$ is a robust and architecture-agnostic default.}
    \label{fig:middle_layer}
\end{figure}

To verify that MSV-Prune is robust to the choice of intermediate vision encoder layer, we conduct layer sweep experiments on two architectures: LLaVA-1.5-7B and Qwen3-VL-8B-Instruct, as shown in Figure~\ref{fig:middle_layer}. For LLaVA-1.5-7B (Figure~\ref{fig:middle_layer}(a)), we vary the middle layer index from 6 to 18 across the 24-layer CLIP ViT backbone under the 25-30$\%$ TFLOPs setting, and observe that both GQA and MMVet scores remain highly stable, fluctuating by less than 0.5 points throughout. For Qwen3-VL-8B-Instruct (Figure~\ref{fig:middle_layer}(b)), a similar sweep from layer 10 to 18 yields equally flat curves on TextVQA and ChartQA, with a variation of less than 0.3 points across all candidate layers. In both cases, performance peaks near the encoder midpoint (layer 12 for LLaVA-1.5-7B and layer 14 for Qwen3-VL-8B-Instruct) and degrades only marginally at the extremes: layers that are too shallow lack sufficient semantic abstraction, while layers too close to the final output sacrifice the fine-grained object-centric details that motivate the use of intermediate features. These results confirm that MSV-Prune is not sensitive to the exact middle layer selection, and that adopting around $\lfloor L/2 \rfloor$ as the default intermediate layer ID provides a reliable and architecture-agnostic choice that requires no dataset-specific tuning.

\begin{figure}
    \centering \includegraphics[width=0.49\textwidth,height=0.25\textwidth]{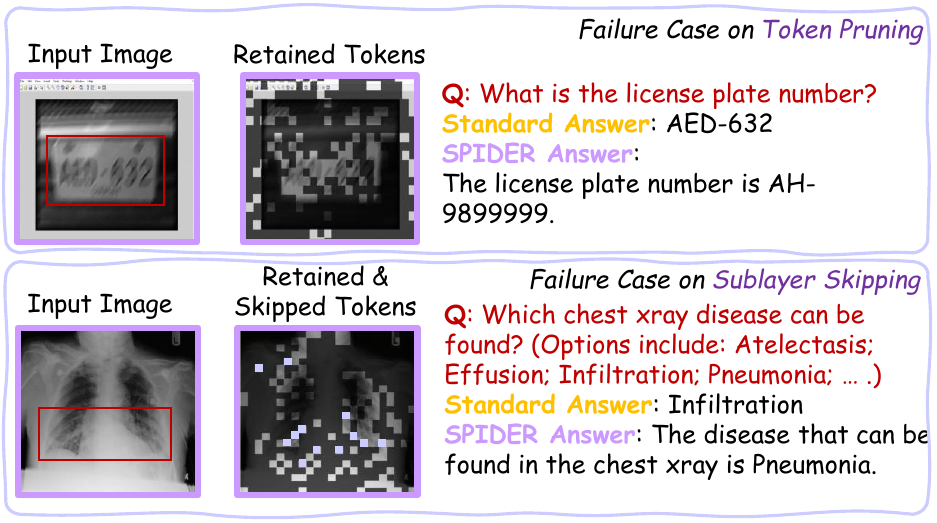}
    \caption{Failure cases: red boxes highlight relevant areas; purple masks show skipped patch tokens during decoding.
}\label{failure}
\end{figure}

\begin{table}[t]
\caption{Stability of the offline SLC policy under the pruned token distribution. We compare the layer-wise SLC ranking computed on dense visual tokens with that recomputed after MSV-Prune. We further report the average accuracy on GQA, MMVet, and POPE when ASL-Skip uses either the original dense-token SLC policy or the recomputed pruned-token SLC policy.}
\centering
\setlength{\tabcolsep}{4pt}
\renewcommand{\arraystretch}{1.15}
\resizebox{0.49\textwidth}{!}{
\begin{tabular}{l|c|cc|c|ccc}
\toprule
\textbf{Model} & \textbf{Retention} & \textbf{Attn} & \textbf{Attn} & \textbf{FFN} & \textbf{Dense-SLC} & \textbf{Pruned-SLC} & $\boldsymbol{\Delta}$ \\
 &  & \textbf{Top-8} & \textbf{Top-16} & \textbf{Top-5} & \textbf{Acc. (\%)} & \textbf{Acc. (\%)} & \textbf{(\%)} \\
\midrule
LLaVA-1.5-7B  & 385/576   & 7/8 & 14/16 & 5/5 &     99.61 & 99.62 & +0.01 \\
LLaVA-NeXT-7B & 1920/2880 & 7/8 & 13/16 & 4/5 &   99.84 & 99.82 & -0.02 \\
\bottomrule
\end{tabular}}
\label{tab:slc_stability}
\end{table}

\subsection{Stability of the offline SLC policy after token pruning.}
Since ASL-Skip uses an offline SLC policy while inference is performed after MSV-Prune, we examine whether token pruning changes the SLC ranking substantially. Table~\ref{tab:slc_stability} shows that the ranking is largely preserved under the pruned token distribution: for both LLaVA-1.5-7B and LLaVA-NeXT-7B, the top-ranked attention and FFN sub-layers exhibit high overlap before and after pruning. Moreover, replacing the original dense-token SLC policy with a pruned-token SLC policy leads to only marginal differences in downstream accuracy. This suggests that the offline SLC captures a stable backbone-level redundancy pattern that remains valid after MSV-Prune.

\subsection{Failure Case Analysis}
Figure~\ref{failure} illustrates two representative failure cases of the SPIDER framework, revealing critical limitations in its token pruning and sublayer skipping mechanisms under challenging real-world conditions. 

First, in the license plate recognition task, motion blur introduces severe degradation in image quality, particularly affecting the clarity of individual digits. As shown in the top row of Figure~\ref{failure}, although the input image contains the correct license plate number "AED-632", the retained tokens after SPIDER's pruning process fail to preserve key character regions. Instead, the model retains ambiguous patches that resemble noise or artifacts, leading it to generate an incorrect answer: "AH-9899999". This demonstrates a fundamental issue: SPIDER’s attention-based token selection mechanism, while effective under clean conditions, lacks robustness against low-level image distortions. It tends to prioritize visually salient but semantically irrelevant regions over degraded yet contextually meaningful ones, resulting in hallucinated outputs. The model fails to maintain sufficient fidelity in regions critical for digit parsing, indicating a need for task-aware attention modulation that can adaptively protect high-stakes visual features during compression.

Second, in the chest X-ray diagnosis example, the failure arises from sub-layer skipping in clinically relevant regions. As shown in the bottom row of Figure~\ref{failure}, the reference answer is ``Infiltration'', whereas SPIDER predicts ``Pneumonia''. The skipped patch tokens, highlighted by the purple masks, overlap with the abnormal region marked by the red box, indicating that critical evidence may be insufficiently updated during decoding. This case suggests that, in medical imaging, indiscriminate compression can distort pathology attribution and lead to clinically misleading predictions. It also highlights the need for domain-aware compression strategies that incorporate anatomical priors or task-specific importance cues to better preserve diagnostically relevant regions.


To mitigate these issues, future work may explore adaptive token retention policies guided by task semantics. For example, integrating a lightweight semantic segmentation module or using anatomical priors in medical contexts could help preserve critical regions even under aggressive pruning.

\section{Conclusion}
We present SPIDER, a training-free framework that jointly performs multi-layer semantic visual token pruning and adaptive sub-layer skipping in MLLMs. By leveraging redundancy patterns across vision encoder layers and differentiating the roles of Attention and FFNs, SPIDER achieves substantial computational savings with minimal accuracy loss. Experiments across diverse MLLMs and benchmarks show large margins over earlier token-pruning and layer-skipping methods, and parity with the strongest recent baselines at matched compute.

\section*{Acknowledgments}
This work was supported by the National Natural Science Foundation of China (No. 62121002, No. 62472396) and the Anhui Provincial Natural
Science Foundation (2508085QF212).


{
    \bibliographystyle{IEEEtran}
    \bibliography{reference.bib}
}

\begin{IEEEbiography}[{\includegraphics[width=1in,height=1.25in,clip,keepaspectratio]{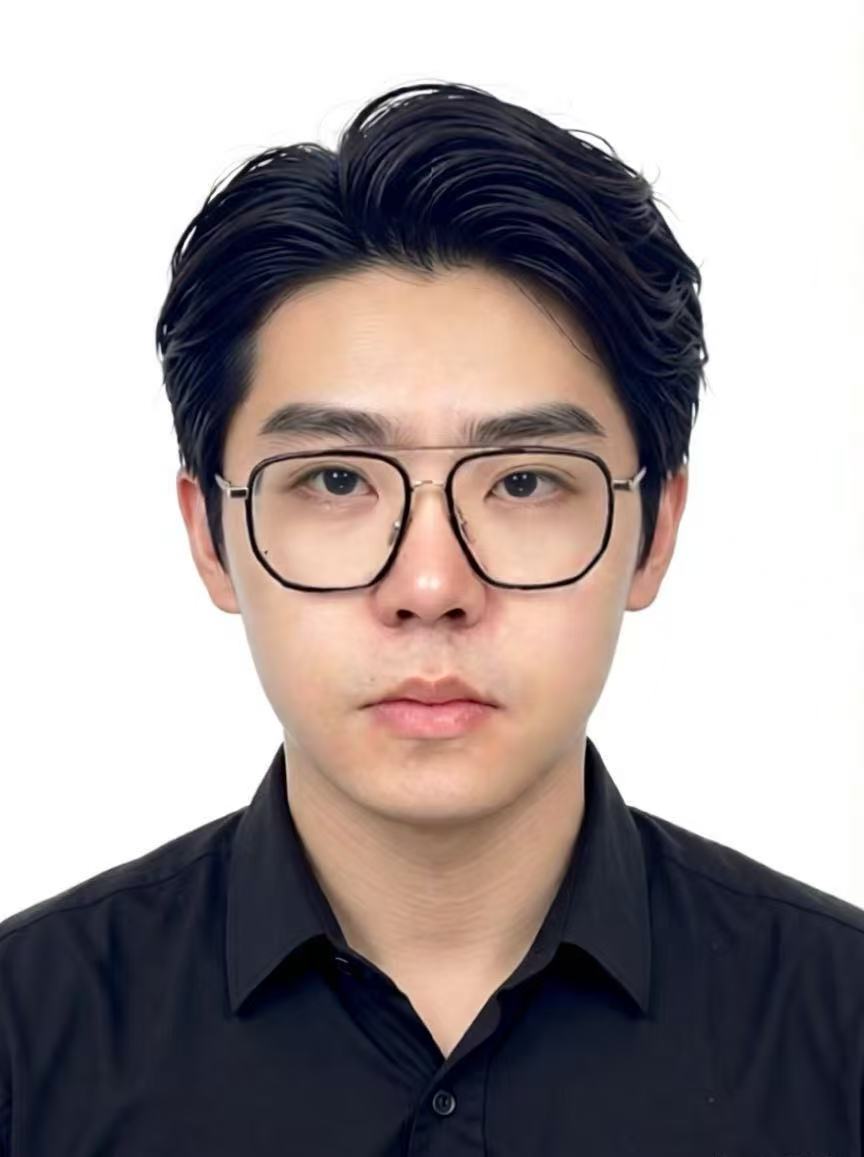}}]{Tianxiang Chen}
received the B.S. degree in mathematics, physics and fundamental sciences from the University of Electronic Science and Technology of China in 2021, and the Ph.D. degree from the School of Cyber Space and Security, University of Science and Technology of China, in 2026. He is currently an algorithm expert at Alibaba Cloud and also a postdoctoral researcher in the joint industry postdoctoral program of Fudan University and Alibaba Cloud. He is a recipient of the Special Award of the President of the Chinese Academy of Sciences. His research interests include Agentic RL, large language models, multi-model large language models and Medical AI.
\end{IEEEbiography}

\begin{IEEEbiography}[{\includegraphics[width=1in,height=1.25in,clip,keepaspectratio]{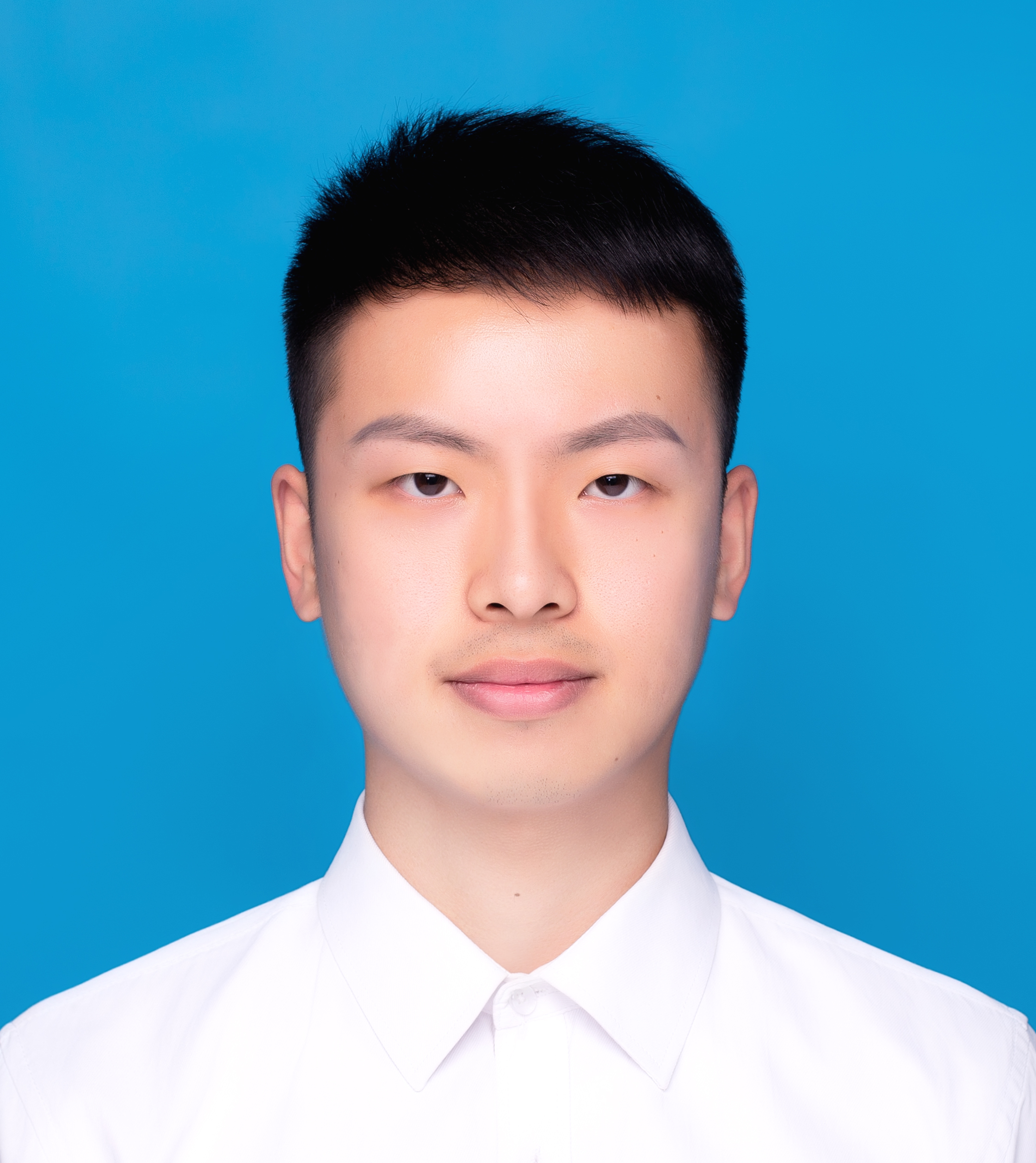}}]{Zhentao Tan}
			received B.S. and Ph.D. degrees from the University of Science and Technology of China in 2017 and 2022, respectively. He is currently a postdoc. in the University of Science and Technology of China and Alibaba Group. His research interests include semantic segmentation, video object segmentation, image synthesis, vision transformers, lightweight models, and large language models.
		\end{IEEEbiography}


\begin{IEEEbiography}[{\includegraphics[width=1in,height=1.25in,clip,keepaspectratio]{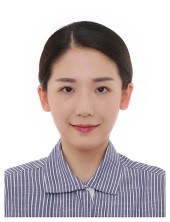}}]{Zi Ye} received a Master’s degree in Applied Statistics from the University of Oxford, UK, in 2010 and a Ph.D. at Universiti Teknikal Malaysia Melaka in 2022. She was a research fellow at Trinity College Dublin, Ireland. Now she is an assistant professor at Maynooth University, Ireland. Her research interests involve Artificial Intelligence $\&$ Machine Learning.
\end{IEEEbiography}

\begin{IEEEbiography}[{\includegraphics[width=1in,height=1.25in,clip,keepaspectratio]{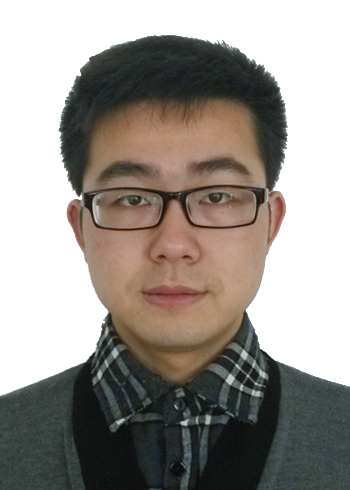}}]{Yue Wu} received a B.E. degree in electronic engineering and a Ph.D. degree in information and communication engineering from the University of Science and Technology of China (USTC), Hefei, China, in 2012 and 2017, respectively. He is currently a senior algorithm expert with the Alibaba Group, in Hangzhou, China. His research interests include multimedia, computer vision, machine learning, and data mining.
		\end{IEEEbiography}

        \begin{IEEEbiography}[{\includegraphics[width=1in,height=1.25in,clip,keepaspectratio]{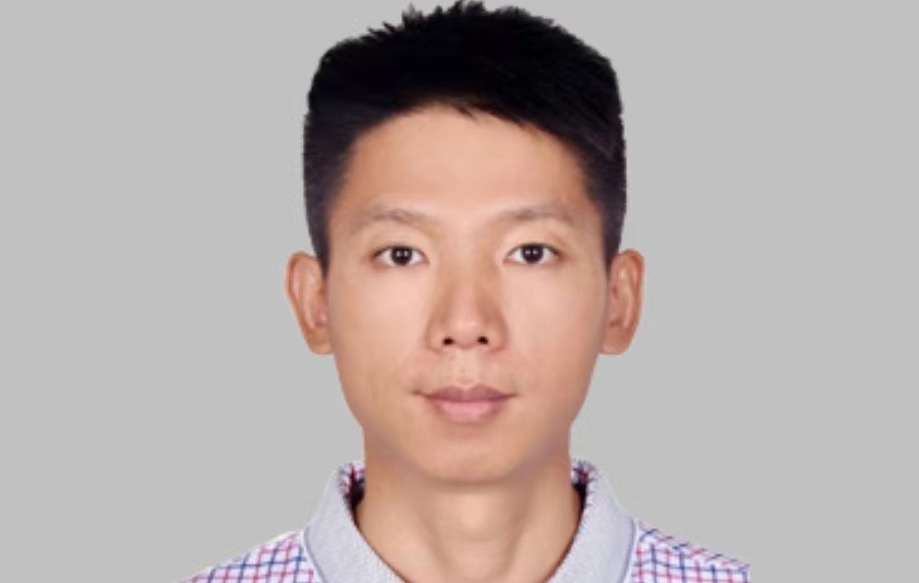}}]{Xiaobing Tu} has extensive experience in system and AI infrastructure at Samsung, Intel, and NVIDIA.
From 2018 to 2021, he focused on model optimization at Alibaba Cloud’s Heterogeneous Computing Team.
From 2021 to 2024, he led Kuaishou’s AI INFRA team, covering generative AI, model optimization, inference, training, and ML Ops.
In 2024, he led SenseTime’s HPC and inference department for large model optimization and domestic computing platforms.
He now leads the Wuying in Alibaba- Cloud LLM Team and holds over 20 patents with several top-conference papers.
		\end{IEEEbiography}

        \begin{IEEEbiography}[{\includegraphics[width=1in,height=1.25in,clip,keepaspectratio]{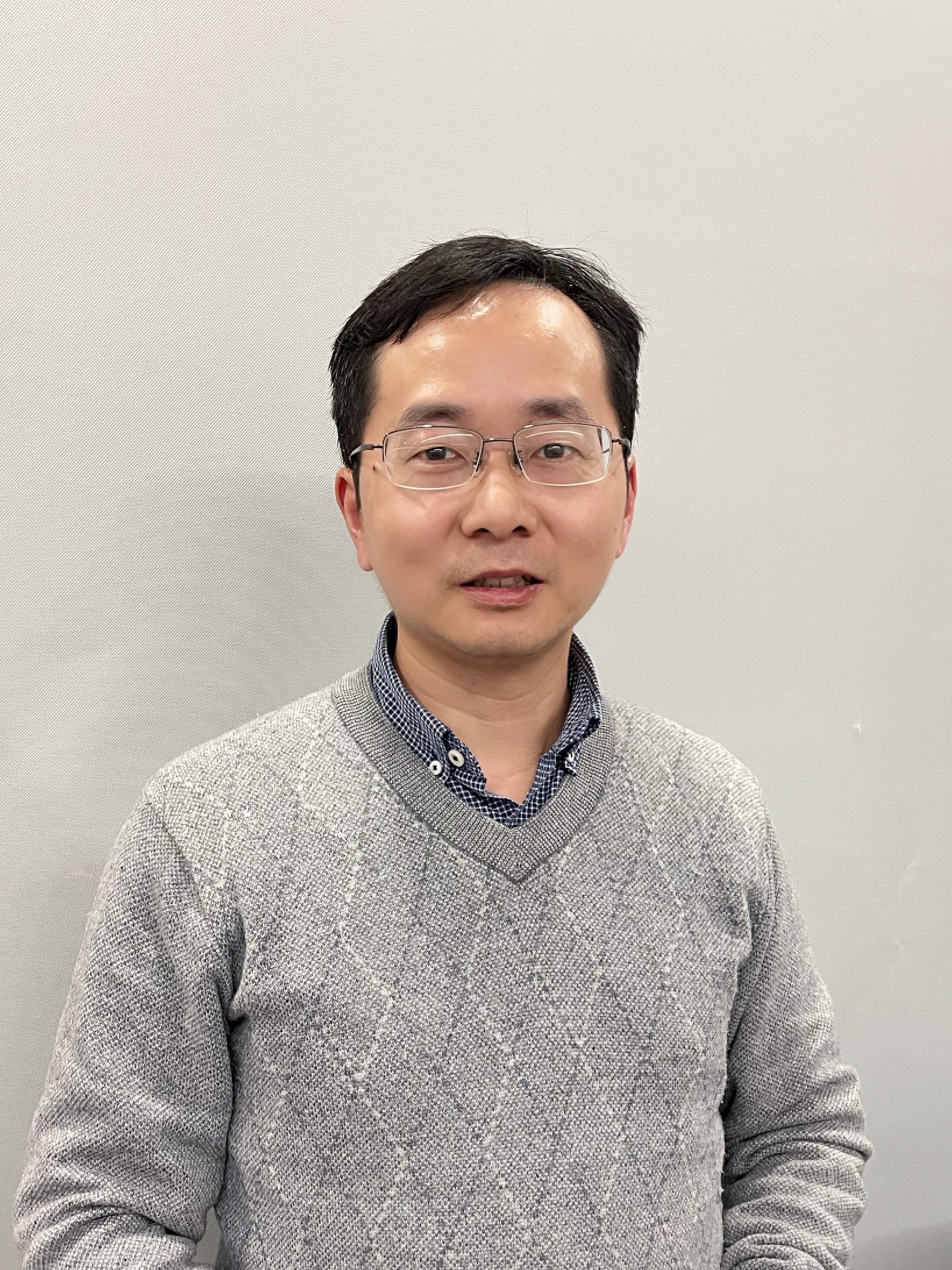}}]{Jinkui Ren}  graduated from the Special Class for the Gifted Young at Shanghai Jiao Tong University, majoring in Electronic Engineering and Communication. After obtaining his master’s degree, he worked successively at Huawei, Intel and Alibaba. He currently serves as Chief Architect of Alibaba’s Wuying and AgentBay product lines. He has extensive expertise in operating systems, virtualization, large language models and related technologies, and holds more than 60 patents.
		\end{IEEEbiography}

        \begin{IEEEbiography}[{\includegraphics[width=1in,height=1.25in,clip,keepaspectratio]{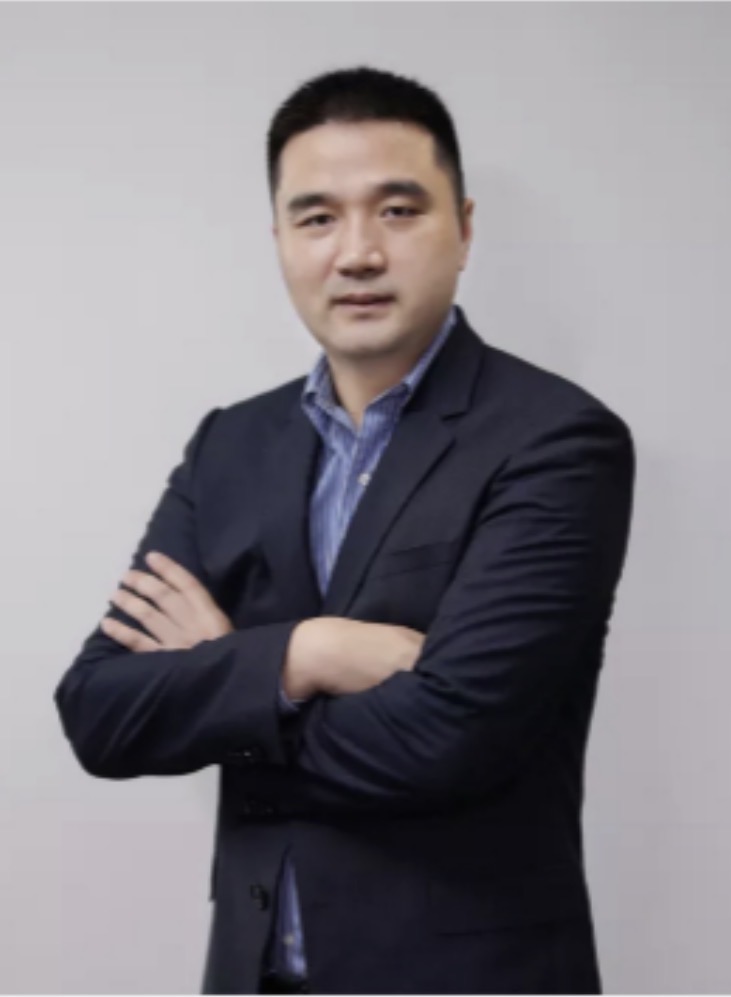}}]{Xiantao Zhang}  holds a Ph.D. $\&$ M.S. in Information Security from Wuhan University. He is President of Alibaba Cloud Wuying Division. Previously at Intel, he led Xen/KVM development and won Intel’s Highest Achievement Award. Since 2014 at Alibaba Cloud, he built the Shenlong Architecture (World Leading Scientific and Technological Achievement Award) and holds 30+ patents. He now drives Wuying’s AI-native strategy, launching AgentBay (China’s first MCP-native cloud service for AI Agents) and AgenticComputer, redefining cloud computing for AI-driven automation.
		\end{IEEEbiography}

  \begin{IEEEbiography}[{\includegraphics[width=1in,height=1.25in,clip,keepaspectratio]{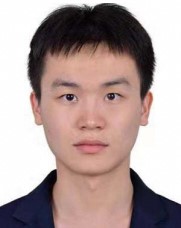}}]{Tao Gong} received the B.E. degree in electronic engineering and the Ph.D. degree in cyber science and technology from University of Science and Technology of China, Hefei, China, in 2016 and 2021, respectively. He is currently an associate research fellow at the University of Science and Technology of China. His research interests include vision and language, object perception, and AI-generated content detection.
		\end{IEEEbiography}

		\begin{IEEEbiography}[{\includegraphics[width=1in,height=1.25in,clip,keepaspectratio]{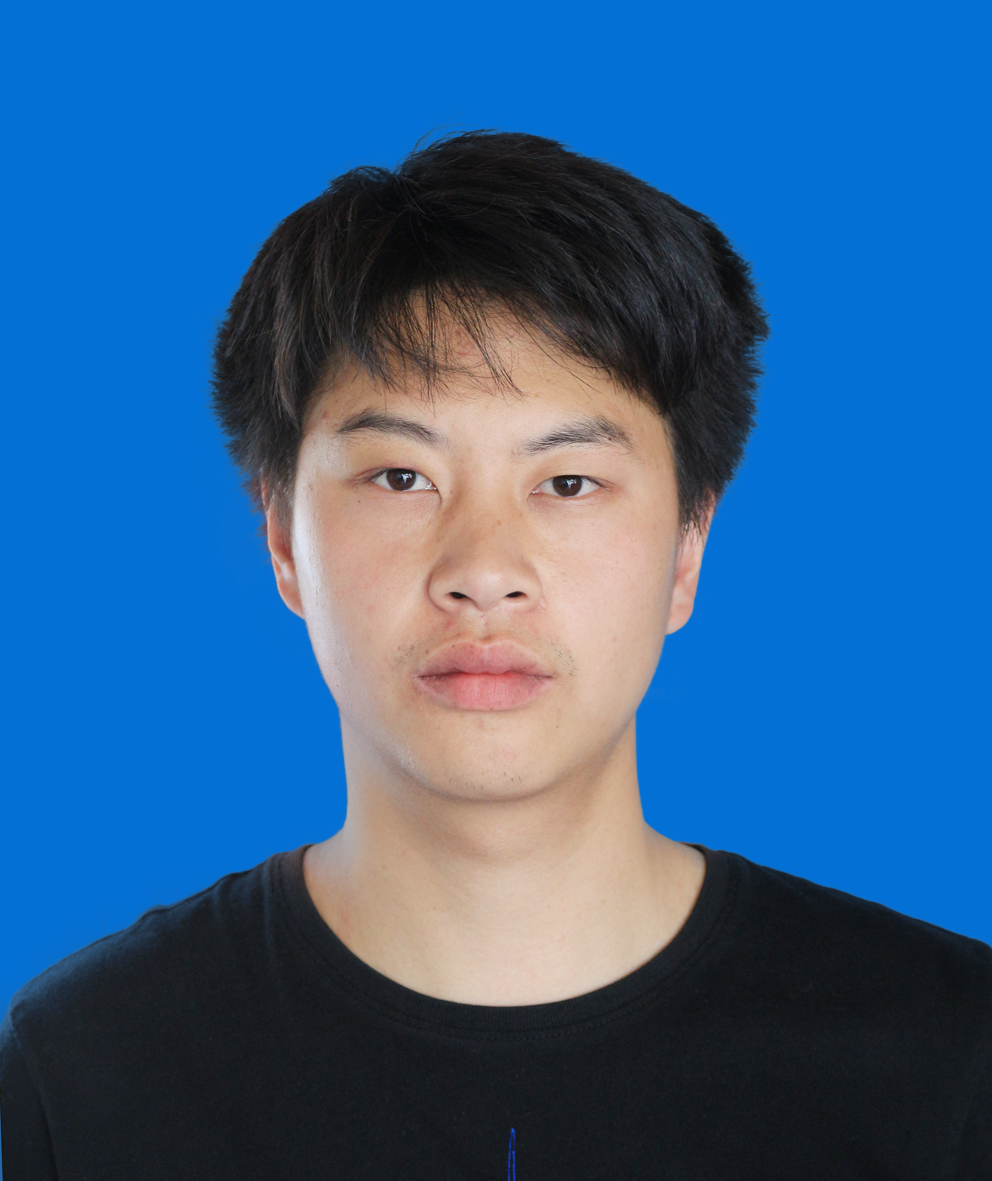}}]{Qi Chu} received a B.S. degree in electronic engineering and a Ph.D. degree in information and communication engineering from University of Science and Technology of China in 2014 and 2019, respectively. Currently, he is an associate research fellow at the University of Science and Technology of China. His research interests include object detection, tracking, image synthesis, and adversarial examples.
		\end{IEEEbiography}

        \begin{IEEEbiography}[{\includegraphics[width=1in,height=1.25in,clip,keepaspectratio]{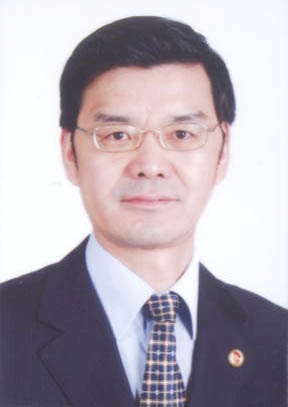}}]{Nenghai Yu} is a full Professor at the University of Science and Technology of China. He is also the director of the Information Processing Center of USTC and deputy director of the academic committee of School of Information Science and Technology. He received the Ph.D. degree from USTC in 2004. He was a visiting scholar at the Institute of Production Technology, Faculty of Engineering, University of Tokyo, in 1999 and performed cooperative research as a senior visiting scholar in the Dept. of Electrical Engineering, Columbia University, from Apr. to Oct. 2008. His research focuses on image processing and video analysis, multimedia communication, media content security, internet information retrieval, data mining and content filtering, network communication, and security.
		\end{IEEEbiography}

        \begin{IEEEbiography}[{\includegraphics[width=1in,height=1.25in,clip,keepaspectratio]{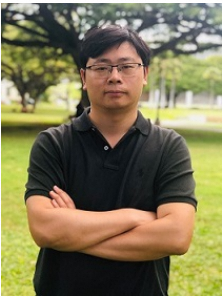}}]{Xipeng Qiu} received the B.Sc. degree and Ph.D.
degrees in computer science from Fudan University,
China in 2001 and 2006, respectively. Currently, he is a professor in the School of Computer Science, Fudan University, China. His research interests include natural language processing and deep learning. He has published more than 60 papers in leading
international journals and conferences, including TACL, ACL, EMNLP,
AAAI, and IJCAI. He received the ACL 2017 Distinguished Paper Award
and the CCL 2019 Best Paper Award. He is the author of FudanNLP, an
open-source Chinese natural language processing toolkit, and the lead
of the FastNLP project. He was selected for the Young Elite
Scientists Sponsorship Program of the China Association for Science
and Technology in 2015, received the First Prize of the Qian
Weichang Award for Young Scholars in Chinese Information Processing
in 2018, and was named an AI 2000 Most Influential Scholar Honorable
Mention in 2020.
		\end{IEEEbiography}

		\begin{IEEEbiography}[{\includegraphics[width=1in,height=1.25in,clip,keepaspectratio]{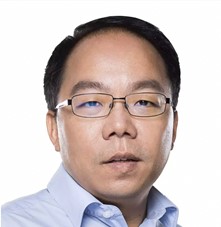}}]{Jieping Ye} received a Ph.D. degree in computer science from the University of Minnesota, Twin Cities, Minnesota, in 2005. He is currently a VP of Alibaba. He is also a professor at the University of Michigan, Ann Arbor, Michigan. His research interests include Big Data, machine learning, and data mining with applications in transportation and biomedicine. He won the NSF CAREER Award, in 2010 and the 2019 Daniel H. Wagner Prize for Excellence in the Practice of Advanced Analytics and Operations. His papers have been selected for the Outstanding Student Paper at ICML, in 2004, the KDD Best Research Paper Runner Up, in 2013, and the KDD Best Student Paper Award, in 2014. He has served as a senior program committee/area chair/program committee vice chair of many conferences, including NIPS, ICML, KDD, IJCAI, ICDM, and SDM. He has served as an associate editor for the Data Mining and Knowledge Discovery and the for IEEE Transactions on Pattern Analysis and Machine Intelligence.
		\end{IEEEbiography}

\newpage

 




\vfill

\end{document}